\documentclass[11pt]{article}

\usepackage[utf8]{inputenc}
\usepackage[T1]{fontenc}
\usepackage[letterpaper,margin=1in]{geometry}
\usepackage{amsmath,amssymb}
\usepackage{graphicx}
\usepackage{booktabs}
\usepackage{array}
\usepackage{tabularx}
\usepackage{colortbl}
\usepackage{url}
\usepackage{multirow}
\usepackage{xcolor}
\usepackage{enumitem}
\usepackage{listings}
\usepackage{placeins}
\usepackage{float}
\usepackage{tikz}
\usepackage[numbers,sort&compress]{natbib}
\usepackage[hidelinks]{hyperref}

\newcolumntype{L}[1]{>{\raggedright\arraybackslash}p{#1}}
\newcolumntype{Y}{>{\raggedright\arraybackslash}X}
\newcommand{\rowtint}{\rowcolor{black!5}}
\newcommand{\thead}[1]{\textbf{#1}}
\newcommand{\medrange}[2]{\mbox{$#1\,{\scriptstyle[#2]}$}}

\newlength{\listlabelwidth}
\newlength{\listleftmargin}
\setlist{leftmargin=\listleftmargin,labelsep=\labelsep,listparindent=0pt,
  topsep=\smallskipamount,partopsep=0pt,parsep=0pt,itemsep=2pt}
\setlist[itemize]{align=right,
  labelindent=0pt,labelwidth=\dimexpr\listleftmargin-\labelsep\relax}
\setlist[enumerate]{align=right,
  labelindent=0pt,labelwidth=\dimexpr\listleftmargin-\labelsep\relax}
\setlist[description]{font=\upshape\bfseries,labelwidth=\listlabelwidth,
  labelindent=\dimexpr\listleftmargin-\labelsep-\listlabelwidth\relax}

\definecolor{cakeCodeBlue}{RGB}{34,82,145}
\definecolor{cakeCodeTeal}{RGB}{20,112,112}
\definecolor{cakeCodeString}{RGB}{154,83,20}
\definecolor{cakeCodeAccent}{RGB}{112,48,140}

\lstdefinestyle{cake}{
  language=Python,
  basicstyle=\ttfamily\scriptsize,
  keywordstyle=\color{cakeCodeBlue}\bfseries,
  keywordstyle=[2]\color{black},
  keywordstyle=[3]\color{cakeCodeTeal},
  commentstyle=\color{black!55},
  stringstyle=\color{cakeCodeString},
  morekeywords=[3]{role,pipeline,barrier,smem,tmem,view,tma_load,mma,wait,
    fence_proxy,range},
  deletekeywords=[2]{range},
  literate={@cake.schedule}{{{\color{cakeCodeAccent}\bfseries @cake.schedule}}}{14},
  backgroundcolor=\color{black!2},
  columns=fullflexible,
  keepspaces=true,
  showstringspaces=false,
  numbers=left,
  numberstyle=\tiny\color{black!55},
  numbersep=6pt,
  frame=tb,
  framerule=0.45pt,
  rulecolor=\color{black!35},
  xleftmargin=1.4em,
  framexleftmargin=1.2em,
  framexrightmargin=0.3em,
  framesep=4pt,
  aboveskip=0.6\baselineskip,
  belowskip=0.4\baselineskip,
  captionpos=b
}
\newcommand{\cake}{\textsc{Cake}}
\newcommand{\cakeir}{\textsc{Cake IR}}
\newcommand{\flashinferpr}[1]{%
  FlashInfer PR~\##1%
  \footnote{\href{https://github.com/flashinfer-ai/flashinfer/pull/#1}%
    {github.com/flashinfer-ai/flashinfer/pull/#1}}%
}

\title{\textbf{CAKE}: \textbf{C}ompiler--\textbf{A}gent Co-Design\\
for Frontier \textbf{K}ernel \textbf{E}volution}
\author{%
Zihao Ye\textsuperscript{1}\thanks{Equal contribution.}\quad
Yingyi Huang\textsuperscript{1}\footnotemark[1]\quad
Hongyi Jin\textsuperscript{2}\quad
Bohan Hou\textsuperscript{2}\\
Junru Shao\textsuperscript{1}\quad
Zhongming Yu\textsuperscript{1}\quad
Jinqi Chen\textsuperscript{1}\quad
Meghan Cowan\textsuperscript{1}\\
Shiyi Cao\textsuperscript{1}\quad
Shanli Xing\textsuperscript{1}\quad
Hanfeng Chen\textsuperscript{1}\quad
Vinod Grover\textsuperscript{1}\\
Tianqi Chen\textsuperscript{1,2}\quad
Luis Ceze\textsuperscript{1}\\[0.6ex]
\small
\textsuperscript{1}NVIDIA\qquad
\textsuperscript{2}Carnegie Mellon University%
}
\date{}

\begin{document}

\maketitle

\begin{abstract}
Work on GPU kernel agents and work on GPU programming languages have advanced
separately, and the gap between them is where expert kernels are lost. Kernel
agents treat the compiler as a fixed black box: they improve proposal, mutation,
and ranking, but the environment returns only compiler errors, correctness
outcomes, and end-to-end timing---signals that never say which program decision
caused a synchronization failure, a hardware-contract violation, or a pipeline
stall, and that cannot grow when a frontier workload exposes a missing
capability. Meanwhile the languages an agent might write are not built for one.
Tile-level DSLs hide the warp specialization, barrier choreography, and
memory-tier placement that separate expert kernels from merely correct ones;
low-level DSLs expose that control but demand a layout calculus that makes agent
errors both likely and hard to localize. We present \cake{}, which co-designs
the two. Agents author \cakeir{}, a typed, hardware-explicit schedule
representation that gives fine-grained control without a layout algebra and
carries enough information for a verifier and cost model to reason about a
program before it is compiled; the harness answers with localized correctness
and performance diagnostics and is itself a target of evolution, so
recurring failures become new verifier rules, IR primitives, cost-model
calibrations, and reusable tactics rather than one-off workarounds. Across
matched implementation-hidden Flash-KMeans clean starts on B200 (three runs per
representation), the best candidate at an 80-million-token budget reaches a
median $1.144\times$ the tuned FlashML baseline with \cakeir{}, versus
$0.928\times$ for direct CUDA/PTX.
Beyond the clean-start benchmark, agent-generated Kimi Delta Attention reaches
a $2.05\times$ geometric-mean speedup over official FlashKDA and is validated
in end-to-end serving. Dispatcher-backed KNN and KMeans families improve
performance by $1.42\times$--$2.12\times$ across more than 400 shapes, and four
kernel changes are available as upstream PRs. \cake{} targets NVIDIA GPUs from
Ampere through Blackwell and separates single-shape evolution from the
generalization and dispatch stage required for library integration.
\end{abstract}

\section{Introduction}
\label{sec:intro}

Coding agents can now write and revise GPU programs in settings where
correctness and performance are measured automatically
\cite{ouyang2024kernelbench,liao2025kernelevolve,dong2026kernelblaster,
accelopt2025,autocomp2025,cheng2026kernelagent,dai2026cudaagent}. Most such
systems still treat the programming environment as a fixed black box: the agent
proposes code, compiles it, runs a numerical test, measures latency, and picks
another edit. The loop works for local tuning, but a crash does not identify the
violated safety or hardware condition, and one latency number does not explain
which program decision limits performance.

Expert kernel programmers work differently. They keep a compact model of the
workload, reason over explicit hardware resources, and carry reusable rules
between kernels. A compiler already holds most of the machinery needed to
externalize that process---structured operation vocabularies, resource models,
legality checks, static analyses, cost models, lowering rules. The question
\cake{} asks is how to make that machinery agent-facing, and how to improve it
when a frontier workload exposes a gap.

\cake{} answers with three commitments. First, agents edit a typed IR rather
than raw CUDA, so hardware decisions are inspectable before code generation.
Second, the compiler returns localized correctness and performance diagnostics
rather than a pass/fail bit, so cheap analysis filters candidates before they
consume GPU time. Third, the harness is itself a target of evolution: repeated
failures become verifier rules, calibration tasks, or new primitives, gated by
corpus tests. \cakeir{} was designed bottom-up through agent-driven abstraction
discovery over a corpus of production kernels, guided by the requirement that
it reproduce the physical schedules and performance of expert-written kernels
(Figure~\ref{fig:ir-evolution}; Appendix~\ref{app:ir-evolution}). The harness is
likewise maintained primarily by agents under human merge gates.

The system supports two entry points, which correspond to the two ways kernel
work actually arrives. It can start from a production kernel in a library such
as FlashInfer or CUTLASS and continue evolving that implementation; or, for a
workload with no mature reference, it can start from a high-level description or
a Triton implementation and let agents choose warp specialization, layout, and
pipeline structure in \cakeir{} while the compiler verifies the program and
generates CUDA.

This paper reports on the system in that second regime as well as the first.
\cake{} targets NVIDIA GPUs from Ampere through Blackwell, and its validated
corpus covers dozens of kernel families---%
attention and linear attention, dense, grouped, and quantized GEMM, MoE
dataflows, normalization, quantization, Top-$K$, KNN and KMeans, and fused graph
kernels. Section~\ref{sec:evaluation} evaluates repeated clean-start evolution
against a tuned baseline, reports frontier-kernel synthesis on model
architectures such as Kimi Delta Attention~\cite{kimi_linear2025}, Gated
DeltaNet~\cite{gated_deltanet2025}, and sparse attention, and reports
reproduction of established kernel families.
Section~\ref{sec:generalization} then treats the step that benchmark numbers
usually skip: turning a kernel tuned at one shape into a dispatcher-backed
family that a serving library can call at any shape.

\begin{figure}[!htbp]
  \centering
  \includegraphics[width=0.99\linewidth]{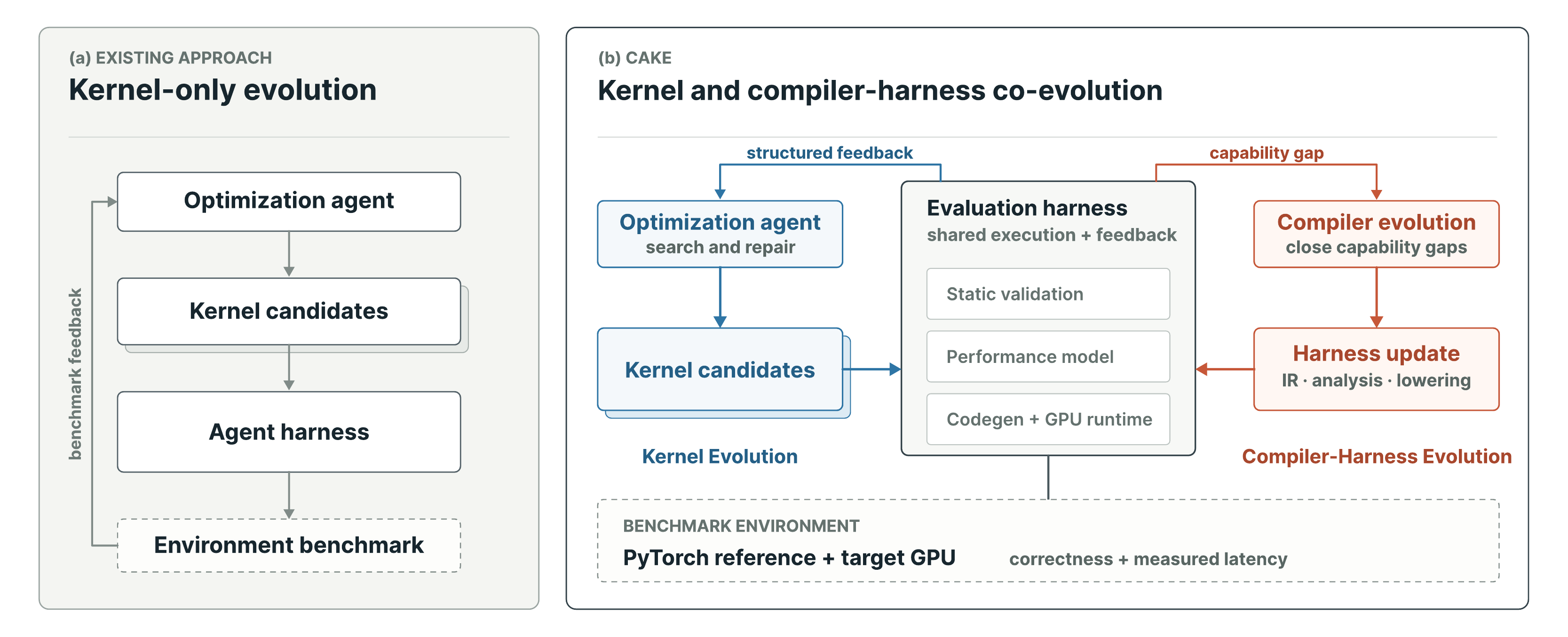}
  \caption{Overview of \cake{}. Kernel evolution consumes structured compiler
  evidence; compiler evolution is the outer loop.}
  \label{fig:overview}
\end{figure}

\FloatBarrier

\section{Program representation}
\label{sec:representation}

\cake{} exposes \cakeir{} as the agent-facing program representation and emits
CUDA/PTX for execution. Structured analysis and performance modeling operate on
\cakeir{}, while conventional sanitizer and profiler feedback comes from the
generated code.

\subsection{Bottom-up IR evolution}
\label{sec:bottom-up-ir}

\begin{figure}[!htbp]
  \centering
  \resizebox{\linewidth}{!}{%
\begin{tikzpicture}[
  font=\sffamily\scriptsize,
  box/.style={rectangle, rounded corners=4pt, draw, line width=0.8pt,
    align=center, inner xsep=7pt, inner ysep=6pt, minimum height=1.15cm},
  flow/.style={->, line width=1.0pt},
  note/.style={font=\sffamily\scriptsize, align=center},
]
\definecolor{irText}{HTML}{1F2933}
\definecolor{irNeutralFill}{HTML}{F7F9FC}
\definecolor{irNeutralBorder}{HTML}{CBD5E1}
\definecolor{irKnowledgeFill}{HTML}{F3F0FF}
\definecolor{irKnowledgeBorder}{HTML}{8B7BD6}
\definecolor{irCandidateFill}{HTML}{FFF4E8}
\definecolor{irCandidateBorder}{HTML}{D9A05B}
\definecolor{irValidationFill}{HTML}{EEF2FF}
\definecolor{irValidationBorder}{HTML}{7687D9}
\definecolor{irSuccessBorder}{HTML}{4F9B65}
\color{irText}

% One external starting point, followed by one repeating three-step loop.
\node[box, fill=irNeutralFill, draw=irNeutralBorder, minimum width=3.0cm]
  (corpus) at (0,0) {\textbf{Production kernels}\\[-1pt]
    \emph{no predefined IR}};
\node[box, fill=irKnowledgeFill, draw=irKnowledgeBorder, minimum width=3.0cm]
  (discover) at (4.2,0) {\textbf{Identify abstraction}\\[-1pt]
    recurring pattern or gap};
\node[box, fill=irCandidateFill, draw=irCandidateBorder, minimum width=3.0cm]
  (revise) at (8.4,0) {\textbf{Revise \cakeir{}}\\[-1pt]
    make the abstraction reusable};
\node[box, fill=irValidationFill, draw=irValidationBorder, minimum width=3.0cm]
  (validate) at (12.6,0) {\textbf{Port + validate}\\[-1pt]
    against the kernel corpus};

\draw[flow, color=irNeutralBorder!80!black]
  (corpus) -- node[note, above] {start} (discover);
\draw[flow, color=irKnowledgeBorder] (discover) -- (revise);
\draw[flow, color=irCandidateBorder] (revise) -- (validate);

% The large return path is the main point of the figure: IR construction never
% terminates at a final box.
\draw[flow, color=irSuccessBorder]
  (validate.south) -- ++(0,-1.25) -| (discover.south);
\node[note, below, color=irSuccessBorder!70!black]
  at (8.4,-1.82) {repeat for the next kernel family or an exposed gap};
\end{tikzpicture}%
}
  \caption{\cakeir{} evolves from kernels rather than a fixed language design.
  The initial corpus enters once; the three-step loop repeats for each new
  kernel family or capability gap.}
  \label{fig:ir-evolution}
\end{figure}
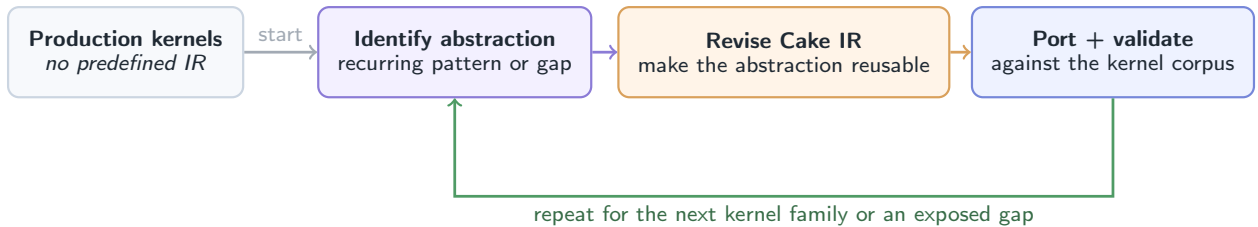

\cake{} did not begin with a predefined \cakeir{} vocabulary. Its starting
material was a corpus of production kernels together with hardware design
principles. From those kernels, agents identify a recurring schedule or a
missing capability, revise the IR and its compiler support, and then port and
validate kernels against the revised system. The next kernel family---or a gap
exposed by validation---starts the same cycle again. This loop both produced
the current IR and continues to grow it (Figure~\ref{fig:ir-evolution});
Appendix~\ref{app:ir-evolution} expands the process.

\FloatBarrier

\subsection{Explicit machine schedules in \cakeir{}}
\label{sec:cake-ir}

\cakeir{} records how the machine should be driven---which warps take which
roles, which buffers are staged how deeply,
which barrier gates which handoff, which instruction form consumes which
operand. The corresponding division of labor is that a schedule states
\emph{what} is to happen and lowering derives \emph{how}: barrier addresses,
phase bits, TMEM offsets, descriptor encodings, and warp identity are all
computed from the declarations rather than written out by the agent.

A program combines explicit operations, declared resources, warp roles, and
grid and pipeline configuration (Figure~\ref{fig:ir-example}). Four properties
do the work. \emph{Type-checked vocabulary}: compute, memory movement,
synchronization, math, and warp control use a fixed IR vocabulary rather than
embedded C or PTX. \emph{Declared resources}: memory regions, synchronization
state, and pipelines are declared once, so the IR knows the shape, dtype, and
lifetime of every buffer. \emph{Explicit roles}: warp groups are named, and
every cross-role handoff is visible rather than an implicit convention.
\emph{Auto-derived metadata}: the mechanical consequences of those declarations
are lowered, not authored.

\begin{figure}[t]
\begin{lstlisting}
@cake.schedule()
def fmha_fwd(lm, Q: LM.tma3d, O: LM.tma2d, seqlen_q: LM.i32):
    # declare resources: named resources, not raw addresses
    pool     = lm.smem(98304)
    smem_q   = pool.view(offset=0, shape=(128,128), dtype=lm.bf16, stage=3)
    tmem_acc = lm.tmem(cols=0, width=128, shape=(128,128), dtype=lm.f32)

    # declare roles: warp groups with assigned work
    load = lm.role(warps=[0])
    mma  = lm.role(warps=[1])
    pipe = lm.pipeline(stages=3)
    q_full = lm.barrier(count=3, prod=[load], cons=[mma],
                        init_count=1, pipeline=pipe)

    with load:
        for stage in lm.range(0, 3):
            smem_q.tma_load(Q, coords=(0,0,stage), stage=stage, barrier=q_full)

    with mma:
        for stage in lm.range(0, 3):
            lm.wait(q_full, stage=stage)
            lm.fence_proxy()
            lm.mma(tmem_acc, smem_q[stage], smem_q[stage], init=(stage == 0))
\end{lstlisting}
\caption{A \cakeir{} schedule fragment. Resources and roles are declared; the
producer--consumer handoff is an explicit barrier; addresses, phases, and warp
identity are derived by lowering.}
\label{fig:ir-example}
\end{figure}

The payoff is that analyses can reason from explicit schedule decisions before
code generation. The harness can therefore tie a finding to the affected
resource, role, or stage rather than returning only a backend error or a hang.
The language catalog, resource model, and design principles are in
Appendix~\ref{app:cake-ir}.

\paragraph{Layout.}
\cake{} deliberately does not make layout a first-class abstraction. Rather
than requiring agents to manipulate a layout algebra, \cakeir{} records storage
and access decisions directly in the schedule. The compiler then checks that
producer and consumer representations are compatible with the target hardware
(Appendix~\ref{app:layout}).

\subsection{Architecture and lowering}
\label{sec:arch}

The same schedule language targets NVIDIA GPUs from Ampere through Blackwell,
so a role--barrier--pipeline schedule is portable in structure while instruction
admission and lowering remain target-specific. \cake{} maps the attached GPU
exactly, reports unsupported targets rather than silently substituting another
architecture, and emits performance estimates only where target-specific
calibration is available. The detailed device matrix and backend notes are
deferred to Appendix~\ref{app:architecture}.

\section{Compiler harness}
\label{sec:harness}

The harness is the agent-facing environment around \cakeir{}. Humans give
high-level descriptions of the intended analyses; agents implement, maintain,
and refine them under validation. During kernel evolution, cheap analyses rank
and filter candidates before they reach expensive GPU runs.

\subsection{Analysis and validation}
\label{sec:analysis}

\paragraph{Program safety and hardware conformance.}
Before compilation, the harness checks the typed schedule for broad classes of
synchronization, memory-safety, data-flow, resource, instruction, and data
representation violations. These checks reject many candidates that are
mathematically plausible but incompatible with the target execution model. A
finding identifies the affected program region and the class of violated
contract, giving the agent a useful repair target through a stable analysis
interface.

\paragraph{Numerical correctness.}
For each workload, we compare the kernel and reference outputs across different
shapes and input distributions. Final acceptance requires end-to-end evaluation
in the corresponding target framework.

\paragraph{Performance modeling.}
A calibrated cost model estimates candidate performance and returns high-level
bottleneck attribution and optimization guidance. The model is used to rank and
filter candidates; on-device measurement and profiling remain the final ground
truth.

Table~\ref{tab:analysis-categories} summarizes the suite by externally visible function
rather than by individual pass or implementation. The important interface is
the contract: blocking checks reject candidates with a localized reason,
reports describe likely performance limits, and hints suggest non-blocking
optimizations.

\begin{table}[H]
  \caption{Analysis and validation categories exposed by the \cake{} harness.
  Categories describe user-visible behavior rather than internal passes.}
  \label{tab:analysis-categories}
  \centering
  \footnotesize
  \setlength{\tabcolsep}{4pt}
  \renewcommand{\arraystretch}{1.06}
  \begin{tabularx}{\linewidth}{@{}L{3.3cm}L{2.1cm}Y@{}}
    \toprule
    \thead{Category} & \thead{Disposition} & \thead{Purpose} \\
    \midrule
    \rowtint
    Program safety & pre-compile gate &
      Identify synchronization, ordering, and memory-use hazards \\
    Hardware conformance & pre-compile gate &
      Enforce supported resource, instruction, and architecture contracts \\
    \rowtint
    Data consistency & pre-compile gate &
      Check data flow and producer--consumer representation compatibility \\
    Schedule semantics & pre-compile gate &
      Check structural invariants of the declared schedule \\
    \rowtint
    Numerical validation & execution gate &
      Compare compiled outputs with an authoritative external reference \\
    Performance analysis & report &
      Estimate cost and identify broad bottleneck classes \\
    \rowtint
    Optimization guidance & hint &
      Suggest promising revisions without blocking compilation \\
    \bottomrule
  \end{tabularx}
\end{table}

\subsection{Compiler evolution}
\label{sec:compiler-evolution}

\cake{} evolves the compiler alongside the kernels. Kernel candidates,
validation results, benchmarks, and failure reports provide evidence for
proposing and validating compiler changes.

\begin{figure}[t]
  \centering
  \includegraphics[width=0.9\linewidth]{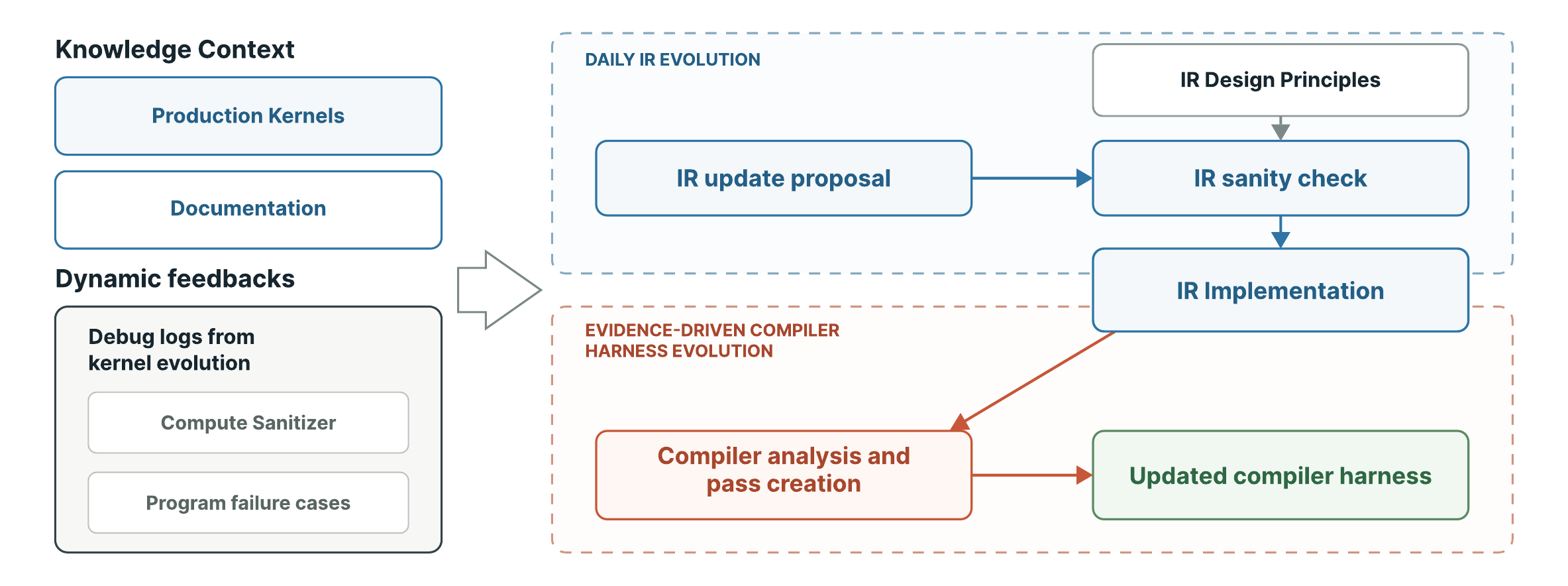}
  \caption{Evidence-driven compiler evolution. Corpus and runtime evidence
  drive validated compiler changes.}
  \label{fig:orchestration}
\end{figure}

Compiler evolution follows two complementary paths
(Figure~\ref{fig:orchestration}).
In the first, agents inspect production kernels and hardware
documentation to find missing Blackwell patterns---new instruction forms,
resource types, descriptor variants, synchronization idioms---and formulate
compiler change proposals. Each proposal is checked against the \cakeir{} design
principles, including performance transparency and verification-friendliness,
before implementation. In the second, agents use feedback from failed
candidates---sanitizer reports, failure cases, correctness mismatches, debugging
logs---and distill recurring or high-cost failure modes into new analyses: an
opaque runtime crash becomes a verifier rule, a repeated illegal lowering
pattern becomes a static check, a systematic misprediction becomes a
calibration target.

The two paths are coupled. New primitives expose additional hardware facts to
the compiler, enabling stronger analysis; new analyses in turn constrain the
design space for future primitives. Compiler changes are test-gated across the
kernel corpus, because a primitive and its analyses must evolve together: syntax
without effects and legality rules makes the IR less analyzable, and a new
verifier rule without corpus validation can reject valid kernels.

\section{Agent workflow}
\label{sec:workflow}

The external workload contract is the stable authority. A run has four stages:
generate structurally distinct \cakeir{} candidates; filter them with IR
construction checks, verifier hard gates, and cost-model ranking before spending
GPU time; evaluate survivors against the external oracle with benchmarking and
profiler evidence; and route the resulting evidence to the candidate, verifier,
cost model, or IR vocabulary according to the diagnosis. The workload contract
fixes the shapes, oracle, tolerances, hardware, and permitted references, while
retained results make decisions auditable and recurring findings reusable.

All reported agent tasks use GPT-5.6-sol \cite{openai2026gpt56} at reasoning
effort \texttt{xhigh}. Holding the model and agent scaffold fixed makes the
comparisons in Section~\ref{sec:evaluation} attributable to the environment
rather than to model capability.

\section{Evaluation}
\label{sec:evaluation}

The evaluation asks three questions: whether the compiler harness can drive
repeated clean-start evolution past a tuned baseline, whether \cake{} can
synthesize frontier kernels without low-level implementation references, and
whether it can reproduce expert kernels against state-of-the-art baselines.

\paragraph{Protocol.}
All measurements use on-GPU correctness checks and CUPTI timing on B200, with
the L2 cache flushed before each timed sample. Each reported candidate is
compiled, checked for correctness, and benchmarked at the listed shape.

\begin{figure}[t]
  \centering
  \includegraphics
    {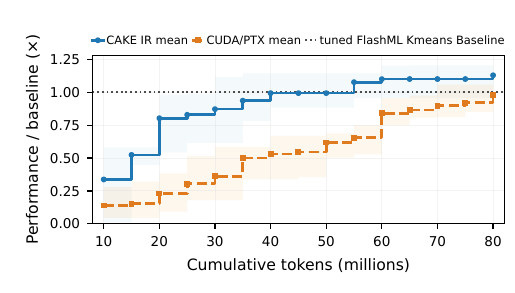}
  \caption{Flash-KMeans fixed-shape clean-start attainment on B200. At each
  5-million-token budget from 10M through 80M, every run contributes its best
  validated speedup so far. Curves show three-run means, bands show run
  minimum--maximum, and the horizontal line is the tuned FlashML K-means
  baseline.}
  \label{fig:clean-start-evolution-token-performance}
\end{figure}

The replicated clean-start runs fix the coding agent and scaffold, model and
reasoning effort, task statement, correctness oracle, benchmark harness, and
single target shape. The Flash-KMeans runs use an isolated B200 clean-start
environment that provides the task specification, correctness oracle, and
benchmark interface while withholding low-level target implementations. The
treatment arm authors typed \cakeir{}; the control arm authors CUDA C++ and
inline PTX directly. We report provider-token consumption and active evolve
time. Table~\ref{tab:kmeans-cleanstart} summarizes three matched runs per arm at
the 80-million-token budget, while
Figure~\ref{fig:clean-start-evolution-token-performance} shows their eligible
performance checkpoints. Summary entries use median
$[\mathrm{min},\mathrm{max}]$; detailed stopping and timing accounting is
retained in the artifact.

\paragraph{Reference access.}
Reference access depends on the question. For clean-start and frontier-kernel
synthesis, agents may inspect the mathematical specification, evaluation
contract, correctness oracle, and high-level code, but not low-level target
implementations such as CUDA, PTX, SASS, or equivalent generated source. Those
references already encode the scheduling decisions the experiment asks the
agent to discover. An external implementation may still be executed through the
benchmark harness as a black-box performance baseline; its internals remain
unavailable to the agent. For known-kernel reproduction, the agent may inspect
the reference. The Flash-KMeans restriction was enforced in isolated
clean-start environments and audited afterward. The direct CUDA/PTX arm changes
the authored representation, not the reference policy: it may write low-level
code but may not inspect an existing target implementation.

\paragraph{Flash-KMeans clean-start workload.}
To test repeatability from an implementation-hidden clean start, we use
Flash-KMeans
\cite{Yangetal2026FlashKMeans}, an exact $k$-means workload motivated by
semantic-aware token permutation in Sparse VideoGen2
\cite{Yangetal2025SparseVideoGen2}. A Lloyd iteration is dominated by two BF16
kernels accounting for more than 95\% of end-to-end time: \texttt{assign},
which computes squared Euclidean distances from each token to all $K$ centroids
and returns an arg-min, and \texttt{centroid\_update}, which reduces tokens
within each cluster into per-cluster sums and counts. The two have different
profiles---\texttt{assign} is a compute-bound BF16 GEMM-and-reduction kernel,
while \texttt{centroid\_update} is bandwidth- and
atomic-contention-sensitive. We focus on \texttt{assign}, which exercises the
tensor-core pipeline and scalar epilogue at $B{=}32$, $N{=}65{,}536$,
$K{=}1024$, and $D{=}128$ with BF16 inputs and FP32 accumulators. All runs use
the same model, oracle, and benchmark. Performance is normalized to the tuned
FlashML KMeans Triton implementation, measured at $0.938$\,ms.

\begin{table}[H]
  \caption{Matched three-run Flash-KMeans clean-start comparison on B200.}
  \label{tab:kmeans-cleanstart}
  \centering
  \footnotesize
  \setlength{\tabcolsep}{3pt}
  \renewcommand{\arraystretch}{1.12}
  \begin{tabular}{@{}lccc@{}}
    \toprule
    \thead{Representation} &
      \thead{\shortstack{Plateau\\by 80M}} &
      \thead{\shortstack{Active\\evolve (h)}} &
      \thead{\shortstack{Best at\\80M}} \\
    \midrule
      \cakeir{}
      & 3/3
      & \medrange{1.89}{1.02,2.33}
      & \medrange{1.144}{1.041,1.205} \\
      Direct CUDA/PTX
      & 0/3
      & \medrange{3.73}{3.59,4.34}
      & \medrange{0.928}{0.852,1.151} \\
    \bottomrule
  \end{tabular}
\end{table}

\paragraph{Trajectory.}
Figure~\ref{fig:clean-start-evolution-token-performance} complements the
terminal summary in Table~\ref{tab:kmeans-cleanstart} by showing when gains
arrive. The \cakeir{} mean crosses the tuned FlashML baseline by 55 million
tokens and continues to improve, while the direct CUDA/PTX mean remains below
baseline at the 80-million-token cutoff.

Across the matched runs, \cakeir{} meets the prespecified plateau criterion in
3/3 runs by 80 million tokens, versus 0/3 for direct CUDA/PTX. Its median best
attainment is $1.144\times$ the tuned FlashML baseline, versus $0.928\times$,
with median active evolve time of 1.89 versus 3.73 hours.

\FloatBarrier
\subsection{Frontier-kernel synthesis}
\label{sec:eval-frontier}

A kernel is frontier here in the operational sense that the agent must discover
its physical schedule without inspecting a low-level target implementation.
Existing implementations may still serve as black-box evaluation baselines.
This is the regime a co-evolving IR should help most, because the search cannot
be anchored to a known-good design, and it is also the regime where the harness
is most exposed, since a missing capability appears as a schedule the agent
cannot express at all.

\paragraph{Emerging model architectures.}
Kimi Delta Attention (KDA) is the clearest case. Official FlashKDA is used only
as a black-box timing baseline; its source and generated code are not provided
to the agent. A FlashKDA-compatible prefill covers fixed, packed-variable, and
tail inputs and reaches a $2.05\times$ geometric-mean speedup over that baseline
across six B200 BF16 shapes. It is bitwise correct on its validation contract
and was verified in end-to-end Kimi-K3 serving under SGLang. The generated CUDA
is available in
\flashinferpr{4262}, so downstream users take on no dependency on \cake{}.
Separate decode paths reach a $1.14\times$ geometric mean over upstream
FlashInfer across 30 public-API shapes (\flashinferpr{4279}). Unlike a GEMM with
an epilogue, KDA contains a recurrent state that must remain live across chunks,
making it a useful test of the schedule representation. Appendix~\ref{app:kda-evolution} reports the corresponding two-phase
source-session trajectory.

Against FlashInfer, the Gated DeltaNet prefill and speculative-decode paths
improve performance while preserving the model's recurrent state. MiniMax
sparse attention further shows that the same representation supports
sparse-attention families across prefill and decode paths. These are dispatch
families rather than single kernels: alternative physical schedules remain
separate \cakeir{} programs behind one logical entry point.

\paragraph{Reference-guided production evolution.}
TinyGEMM provides a complementary production case to the clean-start frontier
experiments above. Starting from FlashInfer's TensorRT-LLM-derived small-$M$
BF16 kernel, the agent produced an adaptive family of shallow and deep
pipelines, including PDL variants and batch sizes below eight.
\flashinferpr{4274} reports an 18--23\% geometric-mean kernel-time reduction
across 35 canonical shapes and a broader regression suite.
Greedy decoding on B200 and GB300 remained bitwise-identical for GPT-OSS-20B
and GPT-OSS-120B; a separate SGLang GPT-OSS-120B experiment reported up to
7.6\% higher output throughput at concurrency 128 on TP1 and differences
within measurement noise on TP4. Appendix~\ref{app:tinygemm-evolution} reports the preceding four-shape search
and targeted small-shape follow-up.

\paragraph{Communication-rich megakernel evolution.}
The original Alpha-MoE implementation \cite{alpha_moe2025} targeted Hopper.
Starting from that implementation, \cake{} agents successfully rewrote its W8A8
fused MoE megakernel for Blackwell. The resulting implementation tests whether
the evolving compiler can express the data exchange around tensor-core
computation, not only the computation itself. It fuses routed gather, two
projections, activation, requantization, and route-weighted output accumulation
into one device program.
Against FlashInfer's TensorRT-LLM-derived pre-routed API, its end-to-end
API-level speedups are $6.204\times$ at $N{=}256$ and $4.025\times$ at
$N{=}512$. A GPU-span remeasurement gives $1.215\times$ and $1.170\times$,
respectively, isolating the improvement in effective GPU execution. The larger
API-level gains additionally reflect fewer scheduling gaps through
launch/schedule fusion and simpler workspace handling: the reference launches
five GPU activities, whereas Alpha-MoE uses an output reset and one megakernel.
The corresponding FlashInfer contribution is \flashinferpr{4287}.
Appendix~\ref{app:alphamoe-evolution} reports the normalized rewrite trajectory,
whose denominator is the initial correct \cake{} checkpoint rather than
TensorRT-LLM.

\FloatBarrier
\subsection{Known-kernel reproduction}
\label{sec:eval-known}

To validate production-quality output, we also target known operator families
with state-of-the-art baselines on modern LLM execution paths: attention
forward/backward and decode, low-precision GEMM, and MQA/MLA logits and decode
kernels. The references come from TensorRT-LLM \cite{tensorrtllm}, CUTLASS
\cite{cutlass}, DeepGEMM \cite{deepgemm2025}, FlashAttention-4
\cite{zadouri2026flashattention4algorithmkernelpipelining}, and FlashInfer
\cite{flashinfer}. These cases test a different question from frontier-kernel
synthesis: when expert structure already exists, can the co-evolved harness help
agents preserve correctness while matching or improving highly optimized
kernels? The comparison itself---the kernel set, the listed reference for each
variant, and the evaluation shape---is fixed, so a rerun changes the values, not
the claim under test. Appendix~\ref{app:known-kernel-details} reports the full
per-kernel matrix, including measured shapes, relative performance, and audited
implementation size (Table~\ref{tab:known-kernel-loc}).

Across the eleven currently measured fixed comparisons, ten entries meet or
exceed the listed reference, and the remaining one reaches 96.5\% of its
reference. The strongest results are the two MQA indexers at roughly
$1.27\times$. All comparisons pass their kernel-specific correctness gates and
use median CUPTI GPU span.

Two effects shape how these results should be read. Variants that land below
their reference generally reflect compiler-integration maturity rather than a
different algorithmic target: where a feature a kernel wants is still being
integrated into the compiler and code generator, the submitted artifact uses
the closest supported strategy. Conversely, the strongest indexer wins are not
faithful transcriptions of the reference kernels. During porting, the agent
explored optimizations absent from the original implementation and retained
variants that passed correctness and benchmarking, so the above-parity entries
reflect search rather than transcription fidelity alone.

The line-count columns are descriptive rather than a cross-language
productivity or readability metric. Every \cakeir{} implementation in the table
is shorter than its audited reference device core, but the languages and
counting scopes differ. The comparison shows only that the evaluated hardware
schedules can be represented compactly in \cakeir{} while retaining the
decisions needed for analysis and lowering.

\subsection{Kernel portfolio}
\label{sec:eval-portfolio}

The preceding evaluations do not convey the breadth of what the harness now
sustains. The validated corpus contains more than 400 static and compile cases
and 399 GPU correctness cases across roughly 28 families, including attention,
dense and sparse GEMM, MoE, quantization, normalization, state-space models,
KNN, and KMeans. It includes architecture-specific paths from Ampere through
Blackwell.

Beyond the frontier kernels of Section~\ref{sec:eval-frontier}, the corpus's
other distinguishing property is composition. Because roles, barriers, and
buffers are declared rather than implied, schedules that would normally be
separate kernels can be expressed as one device program. BatchAttention combines
decode and prefill work, while the Alpha-MoE and mega-MoE families fuse routing,
expert computation, and output accumulation without materializing intermediate
results. The corpus also contains more than 100 TensorRT-LLM ports across
attention, MLA decode, and MoE kernels. The Alpha-MoE W8A8 kernel, originally
written for Hopper, was rewritten by \cake{} agents for Blackwell, illustrating
that schedule structure can survive even when target instructions change.

The four upstream changes cover KDA prefill, KDA decode, TinyGEMM2, and
Alpha-MoE.

\section{From a tuned shape to a library}
\label{sec:generalization}

Everything to this point optimizes a shape. A library takes whatever shape the
caller passes. Closing that gap is not a matter of running the inner loop on
more shapes; it is a separate stage with a different objective, a different
ranking signal, and a different failure mode, and \cake{} treats it as such.

\paragraph{Separate objectives.}
An exact shape gives the inner loop a clean denominator and permits aggressive
specialization. Scoring that loop on broad coverage would weaken this signal.
Generalization therefore begins only after strong per-shape seeds exist and is
scored on dispatcher-inclusive performance over a fixed workload. Incorrect or
slow seeds return to the inner loop rather than being hidden behind routing.

\paragraph{Building and validating portfolios.}
The generalization stage groups measured seeds into shape buckets, produces
specialized or shared variants, and orders their guards behind an explicit
fallback. Tuning may change implementation parameters but not the input shape.
Before reporting an aggregate, validation covers representative and held-out
inputs, boundary and tail cases, overlapping or missing guards, and the
fallback path.

\paragraph{Preventing evaluation leakage.}
The valid shape domain is declared before tuning. Dispatcher predicates may
partition that domain, but they may not introduce convenient new evaluation
rows. Coverage expands through deterministic unseen shards of the same source.
This separation prevents the dispatcher from being tuned on the set used to
claim generalization.

\paragraph{What this looks like in the corpus.}
Several of the kernels in Section~\ref{sec:eval-frontier} are already portfolios
rather than single schedules. KNN build uses coarse outer families with
shape-specific routes within them, whereas KMeans dispatches to a smaller set of
final-route buckets. Attention portfolios similarly route distinct decode and
prefill schedules behind one logical entry point. The route-level breakdown
appears in Appendix~\ref{app:dispatcher-details}
(Figure~\ref{fig:dispatcher-portfolios}).

The classic machine-learning workloads give the cleanest read on what the stage
buys, because their portfolios are large enough that a single shape cannot carry
the result. On GB200, the generalized KNN build, KNN search, and KMeans
implementations contributed to FlashLib achieve dispatcher-inclusive results.
We report $G_{\mathrm{span}}$, the unweighted geometric mean of per-shape
speedups, where each speedup is the reference median CUPTI GPU span divided by
the \cake{} median CUPTI GPU span. The $G_{\mathrm{span}}$ values are
$1.418\times$, $2.116\times$, and $1.803\times$ across 112, 198, and 124 shapes,
respectively, with no incorrect outputs and recall 1.0 for KNN. These
full-portfolio results answer a different question from the three-run,
single-shape Flash-KMeans clean-start cohort in
Table~\ref{tab:kmeans-cleanstart}: the former evaluate complete shape sets after
dispatch, whereas the latter evolves one exact shape. Because the measurements use
different hosts, shape distributions, baselines, and protocols, their difference
is not, by itself, a measured cost of generalization.

One policy limits the practical cost of generalization. The stage reuses a
single physical schedule across as much of the shape domain as it can, and
introduces another only when the domain requires a material schedule change.
Routing complexity must be justified by measured workload gain. Because each
route is a separate \cakeir{} program, alternatives remain independently
analyzable and benchmarkable.

\section{Related work}
\label{sec:related}

\paragraph{GPU programming systems.}
Existing DSLs fall into two camps, both awkward for agent-driven kernel
development. High- and mid-level tile DSLs---Triton \cite{tillet2019triton},
Helion \cite{helion2026}, TileLang \cite{ye2025tilelang}, cuTile
\cite{cutile2026}---hide hardware behind tile abstractions and often automatic
scheduling, but that opacity prevents an agent from expressing the warp
specialization, barrier choreography, and memory-tier placement that separate
expert kernels from merely correct ones. Low-level DSLs such as CuTe DSL
\cite{cutedsl2025} expose hardware control but demand domain-specific expertise
like layout algebra, which is a substantial learning burden and produces brittle
code when layout choices are wrong. Gluon \cite{gluon2026} sits between the two,
reusing Triton's compiler stack while exposing lower-level control over layouts,
memory movement, and asynchrony. \cake{} addresses the gap by co-designing the
IR \emph{with} agents: \cakeir{} gives fine-grained hardware control without
requiring a layout calculus, and it evolves---new primitives when agents find
inexpressible patterns, refined passes when they hit new bug classes,
recalibrated cost models when predictions fail.

\paragraph{Compiler analysis and scheduling.}
TVM \cite{chen2018tvm}, XLA \cite{xla2017}, MLIR \cite{lattner2021mlir},
TensorIR \cite{feng2023tensorir}, Ansor \cite{zheng2020ansor}, and MetaSchedule
\cite{shao2022metaschedule} structure programs for analysis and optimization.
Graphene \cite{hagedorn2023graphene,hagedorn2026time}, Twill
\cite{soi2024twill}, and Tawa \cite{chen2025tawa} model asynchronous GPU
execution, pipelining, or warp specialization. \cake{} shares the principle that
structured programs enable useful analysis, but places compiler findings inside
an agent evolution loop and lets recurring kernel evidence evolve the harness.

\paragraph{Kernel agents.}
KernelBench \cite{ouyang2024kernelbench} established a common evaluation setting
for translating high-level operators into efficient kernels. Several systems
use a human-designed loop in which an LLM revises kernels from compilation,
correctness, or profiling feedback
\cite{accelopt2025,autocomp2025,cheng2026kernelagent}. KernelBlaster
\cite{dong2026kernelblaster} adds a persistent, retrievable CUDA knowledge base
populated from prior optimization experience. KernelEvolve
\cite{liao2025kernelevolve} and EvoEngineer
\cite{guo2025evoengineermasteringautomatedcuda} support evolutionary search over
kernel candidates. AVO \cite{chen2026avo}
replaces fixed mutation and crossover heuristics with autonomous coding agents
as evolutionary variation operators. K-Search \cite{cao2026ksearch} separates
planning from implementation and uses an LLM world model to guide search.
AutoTriton \cite{li2025autotriton} trains a Triton model with supervised
fine-tuning and reinforcement learning. CUDA Agent \cite{dai2026cudaagent}
scales agentic reinforcement learning for CUDA generation and optimization.
These methods evolve the search process, accumulated memory, or model weights
while retaining a chosen DSL and evaluation environment. \cake{}
targets that complementary layer: it changes the representation being searched
and the structured compiler evidence returned to the agent.

\paragraph{Evolving systems.}
FunSearch and AlphaEvolve
\cite{romeraparedes2024funsearch,novikov2025alphaevolve} use population-based
evolutionary search with LLM-generated program mutations. Automated Design of
Agentic Systems \cite{hu2025automated} uses a meta-agent and an archive of prior
discoveries to iteratively propose agent implementations in code. The Darwin
G{\"o}del Machine \cite{zhang2026darwingodelmachineopenended} iteratively modifies coding-agent
implementations, empirically evaluates each variant, and retains variants in an
open-ended archive. Meta-Harness
\cite{lee2026metaharnessendtoendoptimizationmodel} optimizes the code around a
fixed LLM. The self-defining-systems
agenda \cite{anderson2025sds} more broadly considers AI-operated systems that can
change their own mechanisms and abstractions. \cake{} differs in the object
evolved: the coding agent and foundation model remain fixed, while recurring
kernel evidence drives changes to a domain-specific compiler harness: its IR
vocabulary, analyses, and cost calibration, under corpus tests and predefined
merge gates.

\section{Discussion and conclusion}
\label{sec:discussion}

\cake{} now targets NVIDIA architectures from Ampere through Blackwell. The
schedule language, role model, and analysis substrate carry across those
targets, while instruction forms, legality rules, and cost anchors remain
architecture-specific. That cost is the honest measure of transfer. It remains
unmeasured for non-NVIDIA targets, where the backend lowering path would also
have to be rebuilt. Coverage is uneven: most performance evidence is B200, and
the timing model is calibrated only for B200 and H100 and declines to predict
elsewhere. Static analyses and performance models remain intentionally
incomplete---they rank and filter during evolution while GPU execution stays
ground truth
(Appendix~\ref{app:soundness})---and compiler evolution is still human-guided at
merge gates.

\cake{} treats the compiler environment as an evolving collaborator for kernel
agents. \cakeir{} exposes hardware decisions to structured analysis, and the
compiler-evolution loop turns recurring failures into reusable compiler
knowledge. We hope this encourages further work on compiler evolution and
compiler--agent co-design.

{\small
% unsrtnat numbers references in order of first citation; sort&compress keeps
% multi-key groups ascending and collapses runs into ranges.
\bibliographystyle{unsrtnat}
\bibliography{references}

@inproceedings{chen2018tvm,
  author       = {Tianqi Chen and
                  Thierry Moreau and
                  Ziheng Jiang and
                  Lianmin Zheng and
                  Eddie Q. Yan and
                  Haichen Shen and
                  Meghan Cowan and
                  Leyuan Wang and
                  Yuwei Hu and
                  Luis Ceze and
                  Carlos Guestrin and
                  Arvind Krishnamurthy},
  editor       = {Andrea C. Arpaci{-}Dusseau and
                  Geoff Voelker},
  title        = {{TVM:} An Automated End-to-End Optimizing Compiler for Deep Learning},
  booktitle    = {13th {USENIX} Symposium on Operating Systems Design and Implementation,
                  {OSDI} 2018, Carlsbad, CA, USA, October 8-10, 2018},
  pages        = {578--594},
  publisher    = {{USENIX} Association},
  year         = {2018},
  url          = {https://www.usenix.org/conference/osdi18/presentation/chen},
  bibsource    = {dblp computer science bibliography, https://dblp.org}
}

@inproceedings{tillet2019triton,
  title={Triton: An intermediate language and compiler for tiled neural network computations},
  author={Tillet, Philippe and Kung, H. T. and Cox, David},
  booktitle={Workshop on Machine Learning and Programming Languages (MAPL)},
  year={2019},
  url={https://dl.acm.org/doi/10.1145/3315508.3329973}
}

@inproceedings{lattner2021mlir,
  title={{MLIR}: Scaling compiler infrastructure for domain specific computation},
  author={Lattner, Chris and Amini, Mehdi and Bondhugula, Uday and Cohen, Albert and others},
  booktitle={IEEE/ACM International Symposium on Code Generation and Optimization (CGO)},
  year={2021},
  url={https://research.google/pubs/mlir-scaling-compiler-infrastructure-for-domain-specific-computation/}
}

@misc{xla2017,
  title={{XLA}: Optimizing compiler for machine learning},
  author={{XLA Team}},
  year={2017},
  howpublished={Google},
  url={https://openxla.org/xla}
}

@misc{ouyang2024kernelbench,
      title={KernelBench: Can LLMs Write Efficient GPU Kernels?},
      author={Anne Ouyang and Simon Guo and Simran Arora and Alex L. Zhang and William Hu and Christopher Ré and Azalia Mirhoseini},
      year={2025},
      eprint={2502.10517},
      archivePrefix={arXiv},
      primaryClass={cs.LG},
      url={https://arxiv.org/abs/2502.10517},
}

@misc{kimi_linear2025,
  title = {Kimi Linear: An Expressive, Efficient Attention Architecture},
  author = {{Kimi Team}},
  year = {2025},
  eprint = {2510.26692},
  archivePrefix = {arXiv},
  primaryClass = {cs.CL},
  url = {https://arxiv.org/abs/2510.26692}
}

@inproceedings{gated_deltanet2025,
  title = {Gated Delta Networks: Improving Mamba2 with Delta Rule},
  author = {Songlin Yang and Jan Kautz and Ali Hatamizadeh},
  booktitle = {The Thirteenth International Conference on Learning Representations},
  year = {2025},
  url = {https://openreview.net/forum?id=r8H7xhYPwz}
}

@misc{alpha_moe2025,
  author = {{Aleph Alpha}},
  title = {Alpha-MoE: A fused Mixture of Experts megakernel},
  year = {2025},
  publisher = {GitHub},
  journal = {GitHub repository},
  howpublished = {\url{https://github.com/Aleph-Alpha/Alpha-MoE}},
  note = {Software for fused Mixture of Experts kernels compatible with vLLM and SGLang}
}

@misc{bentz2025cutile,
  author = {Bentz, Jonathan and Scudiero, Tony},
  title = {cuTile: Simplify GPU Programming with NVIDIA CUDA Tile in Python},
  year = {2025},
  publisher = {GitHub},
  journal = {GitHub repository},
  howpublished = {\url{https://github.com/NVIDIA/cutile-python}},
  note = {NVIDIA Technical Blog and software repository}
}

@misc{guo2025sonicmoeacceleratingmoeio,
      title={SonicMoE: Accelerating MoE with IO and Tile-aware Optimizations},
      author={Wentao Guo and Mayank Mishra and Xinle Cheng and Ion Stoica and Tri Dao},
      year={2025},
      eprint={2512.14080},
      archivePrefix={arXiv},
      primaryClass={cs.LG},
      url={https://arxiv.org/abs/2512.14080},
}

@article{Yangetal2026FlashKMeans,
  author  = {Yang, Shuo and Xi, Haocheng and Zhao, Yilong and Li, Muyang and Fan, Xiaoze and Zhang, Jintao and Cai, Han and Lin, Yujun and Li, Xiuyu and Keutzer, Kurt and Han, Song and Xu, Chenfeng and Stoica, Ion},
  title   = {Flash-KMeans: Fast and Memory-Efficient Exact K-Means},
  journal = {arXiv preprint arXiv:2603.09229},
  year    = {2026},
  eprint  = {2603.09229},
  archivePrefix = {arXiv},
  primaryClass = {cs.DC},
  url     = {https://arxiv.org/abs/2603.09229},
  doi     = {10.48550/arXiv.2603.09229}
}

@article{Yangetal2025SparseVideoGen2,
  author  = {Yang, Shuo and Xi, Haocheng and Zhao, Yilong and Li, Muyang and Zhang, Jintao and Cai, Han and Lin, Yujun and Li, Xiuyu and Xu, Chenfeng and Chen, Jianfei and Han, Song and Keutzer, Kurt and Stoica, Ion},
  title   = {Sparse VideoGen2: Accelerate Video Generation with Sparse Attention via Semantic-Aware Permutation},
  journal = {arXiv preprint arXiv:2505.18875},
  year    = {2025},
  eprint  = {2505.18875},
  archivePrefix = {arXiv},
  primaryClass = {cs.CV},
  url     = {https://arxiv.org/abs/2505.18875},
  doi     = {10.48550/arXiv.2505.18875}
}

@misc{cutlass,
  title={{CUTLASS}: {CUDA} templates for linear algebra subroutines},
  author={{NVIDIA}},
  howpublished={GitHub},
  url={https://github.com/NVIDIA/cutlass}
}

@misc{tensorrtllm,
  title={{TensorRT-LLM}},
  author={{NVIDIA}},
  howpublished={GitHub},
  url={https://github.com/NVIDIA/TensorRT-LLM}
}

@misc{deepgemm2025,
      title={DeepGEMM: clean and efficient BLAS kernel library on GPU},
      author={Chenggang Zhao and Zhean Xu and Liang Zhao and Jiashi Li and Chenhao Xu and Anyi Xu and Shengyu Liu and Kexing Zhou and Kuai Yu},
      year={2025},
      publisher = {GitHub},
      howpublished = {\url{https://github.com/deepseek-ai/DeepGEMM}},
}

@misc{zadouri2026flashattention4algorithmkernelpipelining,
      title={FlashAttention-4: Algorithm and Kernel Pipelining Co-Design for Asymmetric Hardware Scaling},
      author={Ted Zadouri and Markus Hoehnerbach and Jay Shah and Timmy Liu and Vijay Thakkar and Tri Dao},
      year={2026},
      eprint={2603.05451},
      archivePrefix={arXiv},
      primaryClass={cs.CL},
      url={https://arxiv.org/abs/2603.05451},
}

@inproceedings{
flashinfer,
title={FlashInfer: Efficient and Customizable Attention Engine for {LLM} Inference Serving},
author={Zihao Ye and Lequn Chen and Ruihang Lai and Wuwei Lin and Yineng Zhang and Stephanie Wang and Tianqi Chen and Baris Kasikci and Vinod Grover and Arvind Krishnamurthy and Luis Ceze},
booktitle={Eighth Conference on Machine Learning and Systems},
year={2025},
url={https://openreview.net/forum?id=RXPofAsL8F}
}

@inproceedings{feng2023tensorir,
  title={{TensorIR}: An Abstraction for Automatic Tensorized Program Optimization},
  author={Feng, Siyuan and Hou, Bohan and Jin, Hongyi and Lin, Wuwei and Shao, Junru and Lai, Ruihang and Ye, Zihao and Zheng, Lianmin and Yu, Cody Hao and Yu, Yong and Chen, Tianqi},
  booktitle={International Conference on Architectural Support for Programming Languages and Operating Systems (ASPLOS)},
  year={2023},
  url={https://dl.acm.org/doi/10.1145/3575693.3576933}
}

@inproceedings{zheng2020ansor,
  title={Ansor: Generating High-Performance Tensor Programs for Deep Learning},
  author={Zheng, Lianmin and Jia, Chengfan and Sun, Minmin and Wu, Zhao and Yu, Cody Hao and Haj-Ali, Ameer and Wang, Yida and Yang, Jun and Zhuo, Danyang and Sen, Koushik and Gonzalez, Joseph E. and Stoica, Ion},
  booktitle={USENIX Symposium on Operating Systems Design and Implementation (OSDI)},
  year={2020},
  url={https://www.usenix.org/conference/osdi20/presentation/zheng}
}

@article{shao2022metaschedule,
  title={Tensor Program Optimization with Probabilistic Programs},
  author={Shao, Junru and Zhou, Xiyou and Feng, Siyuan and Hou, Bohan and Lai, Ruihang and Jin, Hongyi and Lin, Wuwei and Masber, Masahiro and Yu, Cody Hao and Chen, Tianqi},
  journal={Advances in Neural Information Processing Systems (NeurIPS)},
  year={2022},
  url={https://dl.acm.org/doi/10.5555/3600270.3602863}
}

@misc{dai2026cudaagent,
      title={CUDA Agent: Large-Scale Agentic RL for High-Performance CUDA Kernel Generation},
      author={Weinan Dai and Hanlin Wu and Qiying Yu and Huan-ang Gao and Jiahao Li and Chengquan Jiang and Weiqiang Lou and Yufan Song and Hongli Yu and Jiaze Chen and Wei-Ying Ma and Ya-Qin Zhang and Jingjing Liu and Mingxuan Wang and Xin Liu and Hao Zhou},
      year={2026},
      eprint={2602.24286},
      archivePrefix={arXiv},
      primaryClass={cs.LG},
      url={https://arxiv.org/abs/2602.24286},
}

@misc{helion2026,
  title={Helion: A High-Level {DSL} for Performant and Portable {ML} Kernels},
  author={{PyTorch Team}},
  year={2025},
  howpublished={PyTorch Blog},
  url={https://pytorch.org/blog/helion/}
}

@misc{ye2025tilelang,
      title={TileLang: A Composable Tiled Programming Model for AI Systems},
      author={Lei Wang and Yu Cheng and Yining Shi and Zhengju Tang and Zhiwen Mo and Wenhao Xie and Lingxiao Ma and Yuqing Xia and Jilong Xue and Fan Yang and Zhi Yang},
      year={2025},
      eprint={2504.17577},
      archivePrefix={arXiv},
      primaryClass={cs.LG},
      url={https://arxiv.org/abs/2504.17577},
}

@misc{cutile2026,
  title={cu{T}ile {P}ython: A Parallel Programming Model for {NVIDIA} {GPUs}},
  author={{NVIDIA}},
  year={2026},
  howpublished={NVIDIA Documentation},
  url={https://docs.nvidia.com/cuda/cutile-python/}
}

@misc{cutedsl2025,
  title={{CuTe DSL}: {P}ython {DSL} for {CUTLASS}},
  author={{NVIDIA}},
  year={2025},
  howpublished={CUTLASS 4 Documentation},
  url={https://docs.nvidia.com/cutlass/latest/media/docs/pythonDSL/cute_dsl.html}
}

@article{cecka2026cute,
  title={{CuTe} Layout Representation and Algebra},
  author={Cecka, Cris},
  journal={arXiv preprint arXiv:2603.02298},
  year={2026},
  eprint={2603.02298},
  archivePrefix={arXiv},
  primaryClass={cs.MS},
  doi={10.48550/arXiv.2603.02298},
  url={https://arxiv.org/abs/2603.02298}
}

@article{zhou2025linear,
  title={Linear Layouts: Robust Code Generation of Efficient Tensor Computation Using {$\mathbb{F}_2$}},
  author={Zhou, Keren and Lezcano, Mario and Goucher, Adam and Rakhmati, Akhmed and Niu, Jeff and Lebar, Justin and Szczerbuk, Pawel and Bell, Peter and Tillet, Phil and Raoux, Thomas and Moudallal, Zahi},
  journal={arXiv preprint arXiv:2505.23819},
  year={2025},
  eprint={2505.23819},
  archivePrefix={arXiv},
  primaryClass={cs.PL},
  doi={10.48550/arXiv.2505.23819},
  url={https://arxiv.org/abs/2505.23819}
}

@article{hou2026axe,
  title={{Axe}: A Simple Unified Layout Abstraction for Machine Learning Compilers},
  author={Hou, Bohan and Jin, Hongyi and Wang, Guanjie and Chen, Jinqi and Cai, Yaxing and Yang, Lijie and Ye, Zihao and Ding, Yaoyao and Lai, Ruihang and Chen, Tianqi},
  journal={arXiv preprint arXiv:2601.19092},
  year={2026},
  eprint={2601.19092},
  archivePrefix={arXiv},
  primaryClass={cs.DC},
  doi={10.48550/arXiv.2601.19092},
  url={https://arxiv.org/abs/2601.19092}
}

@techreport{anderson2025sds,
  title={Self-Defining Systems},
  author={Anderson, Thomas and Mahajan, Ratul and Peter, Simon and Zettlemoyer, Luke},
  institution={Paul G. Allen School of Computer Science \& Engineering, University of Washington},
  year={2025},
  url={https://foci.uw.edu/papers/whitepaper2025-sds.pdf}
}

@article{chen2026avo,
  title={{AVO}: Agentic Variation Operators for Autonomous Evolutionary Search},
  author={Chen, Terry and Ye, Zhifan and Xu, Bing and Ye, Zihao and Liu, Timmy and Hassani, Ali and Chen, Tianqi and Kerr, Andrew and Wu, Haicheng and others},
  journal={arXiv preprint arXiv:2603.24517},
  year={2026},
  url={https://arxiv.org/abs/2603.24517}
}

@article{cao2026ksearch,
  title={{K-Search}: {LLM} Kernel Generation via Co-Evolving Intrinsic World Model},
  author={Cao, Shiyi and Mao, Ziming and Gonzalez, Joseph E. and Stoica, Ion},
  journal={arXiv preprint arXiv:2602.19128},
  year={2026},
  url={https://arxiv.org/abd/2602.19128}
}

@article{novikov2025alphaevolve,
  title={{AlphaEvolve}: A Coding Agent for Scientific and Algorithmic Discovery},
  author={Novikov, Alexander and Vu, Ngan and Eisenberger, Marvin and others},
  journal={arXiv preprint arXiv:2506.13131},
  year={2025},
  url={https://arxiv.org/abs/2512.23236}
}

@article{romeraparedes2024funsearch,
  title={Mathematical Discoveries from Program Search with Large Language Models},
  author={Romera-Paredes, Bernardino and Barekatain, Mohammadamin and others},
  journal={Nature},
  volume={625},
  pages={468--475},
  year={2023},
  url={https://www.nature.com/articles/s41586-023-06924-6}
}

@misc{liao2025kernelevolve,
      title={KernelEvolve: Scaling Agentic Kernel Coding for Heterogeneous AI Accelerators at Meta},
      author={Gang Liao and Hongsen Qin and Ying Wang and Alicia Golden and Michael Kuchnik and Yavuz Yetim and Jia Jiunn Ang and Chunli Fu and Yihan He and Samuel Hsia and Zewei Jiang and Dianshi Li and Uladzimir Pashkevich and Varna Puvvada and Feng Shi and Matt Steiner and Ruichao Xiao and Nathan Yan and Xiayu Yu and Zhou Fang and Roman Levenstein and Kunming Ho and Haishan Zhu and Alec Hammond and Richard Li and Ajit Mathews and Kaustubh Gondkar and Abdul Zainul-Abedin and Ketan Singh and Hongtao Yu and Wenyuan Chi and Barney Huang and Sean Zhang and Noah Weller and Zach Marine and Wyatt Cook and Carole-Jean Wu and Gaoxiang Liu},
      year={2026},
      eprint={2512.23236},
      archivePrefix={arXiv},
      primaryClass={cs.LG},
      url={https://arxiv.org/abs/2512.23236},
}

@misc{tao2026digestion,
  author = {Tao, Terence},
  title = {On the era of proof abundance: generation, verification, and digestion},
  year = {2026},
  howpublished = {Mastodon thread, \url{https://mathstodon.xyz/@tao/116477351524980995}},
  note = {Accessed 2026-04-30}
}

@article{dong2026kernelblaster,
  title={{KernelBlaster}: Continual Cross-Task {CUDA} Optimization via Memory-Augmented In-Context Reinforcement Learning},
  author={Dong, Kris Shengjun and Modi, Sahil and Nikiforov, Dima and Damani, Sana and Lin, Edward and Hari, Siva Kumar Sastry and Kozyrakis, Christos},
  journal={arXiv preprint arXiv:2602.14293},
  year={2026},
  url={https://arxiv.org/abs/2602.14293}
}

@misc{accelopt2025,
      title={AccelOpt: A Self-Improving LLM Agentic System for AI Accelerator Kernel Optimization},
      author={Genghan Zhang and Shaowei Zhu and Anjiang Wei and Zhenyu Song and Allen Nie and Zhen Jia and Nandita Vijaykumar and Yida Wang and Kunle Olukotun},
      year={2026},
      eprint={2511.15915},
      archivePrefix={arXiv},
      primaryClass={cs.LG},
      url={https://arxiv.org/abs/2511.15915},
}

@misc{autocomp2025,
      title={Autocomp: A Powerful and Portable Code Optimizer for Tensor Accelerators},
      author={Charles Hong and Sahil Bhatia and Alvin Cheung and Yakun Sophia Shao},
      year={2025},
      eprint={2505.18574},
      archivePrefix={arXiv},
      primaryClass={cs.PL},
      url={https://arxiv.org/abs/2505.18574},
}

@misc{cheng2026kernelagent,
  author={Cheng, Kaiming and Wang, Laura and Khuu, Jack and Saroufim, Mark and Chi, Wenyuan and Wang, Jiannan and Isaacson, Joe},
  title={{KernelAgent}: Hardware-Guided {GPU} Kernel Optimization via Multi-Agent Orchestration},
  year={2026},
  howpublished={PyTorch Blog},
  url={https://pytorch.org/blog/kernelagent-hardware-guided-gpu-kernel-optimization-via-multi-agent-orchestration/}
}

@inproceedings{hagedorn2023graphene,
  author={Hagedorn, Bastian and Fan, Bin and Chen, Hanfeng and Cecka, Cris and Garland, Michael and Grover, Vinod},
  title={Graphene: An {IR} for Optimized Tensor Computations on {GPUs}},
  booktitle={Proceedings of the 28th ACM International Conference on Architectural Support for Programming Languages and Operating Systems (ASPLOS)},
  year={2023},
  doi={10.1145/3582016.3582018}
}

@inproceedings{hagedorn2026time,
  author={Hagedorn, Bastian and Grover, Vinod},
  title={It's about Time: Temporal Abstractions for Asynchronous {GPU} Tensor Computations},
  booktitle={Proceedings of the 35th ACM SIGPLAN International Conference on Compiler Construction (CC)},
  year={2026},
  doi={10.1145/3771775.3786277}
}

@article{soi2024twill,
  author={Soi, Rupanshu and Yadav, Rohan and Kjolstad, Fredrik and Aiken, Alex and Mehri Dehnavi, Maryam and Garland, Michael and Bauer, Michael},
  title={Optimal Software Pipelining and Warp Specialization for Tensor Core {GPUs}},
  journal={arXiv preprint arXiv:2512.18134},
  year={2024},
  url={https://arxiv.org/abs/2512.18134}
}

@misc{chen2025tawa,
      title={Tawa: Automatic Warp Specialization for Modern GPUs with Asynchronous References},
      author={Hongzheng Chen and Bin Fan and Alexander Collins and Bastian Hagedorn and Evghenii Gaburov and Masahiro Masuda and Matthew Brookhart and Chris Sullivan and Jason Knight and Zhiru Zhang and Vinod Grover},
      year={2025},
      eprint={2510.14719},
      archivePrefix={arXiv},
      primaryClass={cs.LG},
      url={https://arxiv.org/abs/2510.14719},
}

@misc{openai2026gpt56,
  author={OpenAI},
  title={{GPT-5.6 Sol Model}},
  year={2026},
  howpublished={OpenAI API Documentation},
  url={https://developers.openai.com/api/docs/models/gpt-5.6-sol},
  note={Accessed: 2026-08-09}
}

@misc{gluon2026,
  title={Gluon: A Lower-Level {GPU} Programming Language on the {Triton} Compiler Stack},
  author={{Triton Contributors}},
  year={2026},
  howpublished={Triton Documentation and Tutorials},
  url={https://github.com/triton-lang/triton/tree/main/python/tutorials/gluon},
  note={Accessed: 2026-07-27}
}

@misc{zhang2026darwingodelmachineopenended,
      title={Darwin Godel Machine: Open-Ended Evolution of Self-Improving Agents}, 
      author={Jenny Zhang and Shengran Hu and Cong Lu and Robert Lange and Jeff Clune},
      year={2026},
      eprint={2505.22954},
      archivePrefix={arXiv},
      primaryClass={cs.AI},
      url={https://arxiv.org/abs/2505.22954}, 
}

@inproceedings{hu2025automated,
  title={Automated design of agentic systems},
  author={Hu, Shengran and Lu, Cong and Clune, Jeff},
  booktitle={International Conference on Learning Representations},
  volume={2025},
  pages={21344--21377},
  year={2025}
}

@misc{lee2026metaharnessendtoendoptimizationmodel,
      title={Meta-Harness: End-to-End Optimization of Model Harnesses}, 
      author={Yoonho Lee and Roshen Nair and Qizheng Zhang and Kangwook Lee and Omar Khattab and Chelsea Finn},
      year={2026},
      eprint={2603.28052},
      archivePrefix={arXiv},
      primaryClass={cs.AI},
      url={https://arxiv.org/abs/2603.28052}, 
}

@misc{guo2025evoengineermasteringautomatedcuda,
      title={EvoEngineer: Mastering Automated CUDA Kernel Code Evolution with Large Language Models}, 
      author={Ping Guo and Chenyu Zhu and Siyuan Chen and Fei Liu and Xi Lin and Zhichao Lu and Qingfu Zhang},
      year={2025},
      eprint={2510.03760},
      archivePrefix={arXiv},
      primaryClass={cs.LG},
      url={https://arxiv.org/abs/2510.03760}, 
}

@article{li2025autotriton,
  title={Autotriton: Automatic triton programming with reinforcement learning in llms},
  author={Li, Shangzhan and Wang, Zefan and He, Ye and Li, Yuxuan and Shi, Qi and Li, Jianling and Hu, Yonggang and Che, Wanxiang and Han, Xu and Liu, Zhiyuan and others},
  journal={arXiv preprint arXiv:2507.05687},
  year={2025}
}
}

\appendix

\section{IR abstraction evolution process}
\label{app:ir-evolution}

Figure~\ref{fig:ir-evolution} summarizes the feedback loop that produced
\cakeir{} without a predefined language vocabulary. This appendix expands that
agent-driven abstraction-discovery process.

\begin{enumerate}
  \item \textbf{Corpus collection.} High-quality CUDA kernels are collected from
  production libraries \cite{alpha_moe2025,cutlass,bentz2025cutile,
  deepgemm2025,zadouri2026flashattention4algorithmkernelpipelining,
  Yangetal2026FlashKMeans,flashinfer,guo2025sonicmoeacceleratingmoeio,
  ye2025tilelang}. Kernels originally written in other DSLs are first translated
  to CUDA with inline PTX by coding agents.
  \item \textbf{Abstraction extraction.} Agents analyze the corpus and summarize
  recurring patterns---barrier choreography, pipeline staging, warp-role
  partitioning, TMA descriptor setup, TMEM accumulator lifecycles---into
  candidate abstractions.
  \item \textbf{Hardware-informed design.} Human expertise biases the
  abstraction toward the Blackwell programming model: TMEM as a first-class
  resource, warp specialization as the primary parallelism idiom, asynchronous
  barriers as the synchronization primitive, cluster-scoped operations for
  multi-SM coordination.
  \item \textbf{Principle-driven iteration.} Each candidate is validated against
  the eight principles in Appendix~\ref{app:cake-ir}; violations are refined or
  rejected.
  \item \textbf{Port-driven expansion.} New kernels are continuously ported.
  Each port either succeeds, validating the abstraction, or reveals a gap,
  triggering a proposal to extend the IR or its lowering.
\end{enumerate}

The process is ongoing: every new kernel family stress-tests the IR and drives
further evolution.

\section{\cakeir{} design and language constructs}
\label{app:cake-ir}

\subsection{Design principles}

\cakeir{} is guided by eight design goals.

\begin{description}
  \item[P1 Ergonomic.] Keep the editing model familiar to NumPy/PyTorch users
  and avoid unnecessary destination-passing or grid bookkeeping.
  \item[P2 Performance-transparent.] Keep performance-relevant hardware
  decisions visible and lowering behavior inspectable.
  \item[P3 Canonical.] Prefer one canonical form for each operation over
  equivalent alternative spellings.
  \item[P4 Statically type-checked.] Use typing rules to constrain operation
  lowering and reject ill-typed programs during construction.
  \item[P5 Analysis-friendly.] Expose the information required by the supported
  static analyses.
  \item[P6 Test-gated.] Evaluate IR changes against the kernel-matrix tests for
  static analysis and compilation.
  \item[P7 Analysis-consistent.] Accompany changes to the IR data model with
  corresponding analysis updates.
  \item[P8 Hardware-grounded.] Document the intended hardware behavior of each
  operation.
\end{description}

P5 and P7 keep new primitives amenable to static analysis; P6 guards against
regressions as the IR evolves; P2 and P8 keep the mapping to hardware legible to
humans and agents alike. This serves \cake{}'s goal of inspectable, reusable
artifacts---a framing related to Tao's observation, on AI-generated
mathematical proofs, that
abundance of generation shifts the bottleneck from producing artifacts to
verifying and understanding them \cite{tao2026digestion}.

\subsection{Capability classes}

\cakeir{} groups its operation vocabulary into four broad classes.

\begin{description}
  \item[Compute.] Matrix, elementwise, and reduction operations across the
  precision modes supported by the target.
  \item[Memory movement.] Explicit transfers among global and on-chip memory
  tiers, including asynchronous and collective transfer patterns.
  \item[Synchronization.] Ordering and coordination across roles, pipeline
  stages, and hardware scopes.
  \item[Control and scheduling.] Warp-role assignment, pipeline management,
  persistent execution, and multi-block coordination.
\end{description}

\subsection{Declarative resource model}

Programs declare five broad kinds of resources: shared-memory regions,
tensor-memory regions, synchronization objects, warp roles, and pipelines.
Their declarations record the information needed to lower and analyze the
schedule, including data shape, ownership, and lifetime.

Hardware-sensitive choices remain visible in the program, while purely
mechanical metadata is derived during lowering. This balance keeps generated
code inspectable and gives the analyses concrete schedule decisions without
requiring the agent to manipulate raw addresses.

\subsection{Layout verification}
\label{app:layout}

Recent compiler systems make layout an explicit algebraic object: CuTe's layout
algebra, Triton's linear layouts over $\mathbb{F}_2$, and Axe's named-axis
abstraction \cite{cecka2026cute,zhou2025linear,hou2026axe}. \cake{} takes the
opposite position. Layout is not a first-class citizen of \cakeir{}; the agent
writes down the concrete commitments---an SMEM view offset, an operand byte
offset, a TMEM column range, a swizzle tag, a TMA descriptor coordinate---and
the compiler carries the burden of deciding whether those commitments are legal.

The compiler checks that these commitments remain mutually consistent along the
program's data flow and satisfy the target instruction and resource contracts.
This catches representation errors before execution while keeping the agent's
editing surface concrete.

Diagnostics point to the relevant IR decision and a broad mismatch category.
This interface preserves the co-evolution property: new primitives can extend
verification coverage without requiring agents to manipulate a separate layout
language. This paper therefore characterizes the verifier through its contract
and coverage rather than its internal representation or decision procedure.

\subsection{Architecture and backend coverage}
\label{app:architecture}

Table~\ref{tab:arch} records the target-specific coverage behind the summary in
Section~\ref{sec:arch}.

\begin{table}[H]
  \caption{Detailed architecture coverage.}
  \label{tab:arch}
  \centering
  \footnotesize
  \setlength{\tabcolsep}{4pt}
  \renewcommand{\arraystretch}{1.06}
  \begin{tabularx}{\linewidth}{@{}L{1.6cm}L{4.4cm}Y@{}}
    \toprule
    \thead{Target} & \thead{SKU} &
    \thead{Tensor-core path and notable features} \\
    \midrule
    \rowtint
    \texttt{sm\_80} & A100 &
      \texttt{mma.sync}, \texttt{ldmatrix}, \texttt{cp.async}; no TMA,
      clusters, or TMEM \\
    \texttt{sm\_89} & L40S / RTX 6000 Ada &
      \texttt{mma.sync}, \texttt{ldmatrix}, \texttt{cp.async}; FP8 tensor
      cores, no TMA or TMEM \\
    \rowtint
    \texttt{sm\_90a} & H100 / H200 &
      WGMMA + \texttt{mma.sync}; TMA, thread-block clusters, async barriers,
      DSM \\
    \texttt{sm\_100a} & B200 &
      \texttt{tcgen05.mma} + TMEM; 2-CTA MMA (\texttt{cta\_group::2}),
      \texttt{tcgen05.\{ld,cp,shift\}} \\
    \rowtint
    \texttt{sm\_103a} & B300 &
      adds \texttt{tcgen05.ld.red}, $K{=}96$ block-scaled MMA \\
    \texttt{sm\_120a} & RTX 5090 / RTX PRO 6000 &
      \texttt{mma.sync} + \texttt{ldmatrix}, TMA, clusters, DSM; no
      \texttt{tcgen05}/TMEM \\
    \rowtint
    \texttt{sm\_121a} & DGX Spark (GB10) &
      same instruction surface as \texttt{sm\_120a}, separate cubin, different
      SFU rates \\
    \bottomrule
  \end{tabularx}
\end{table}

The compiler requires an exact target match and reports missing device or
toolchain support rather than stepping a schedule down to another architecture.
Timing-model coverage is separately evidence-gated: B200 is the measured
baseline, H100 is calibrated independently, and other targets report a coverage
limitation instead of inheriting estimates.

After static checks, \cakeir{} lowers deterministically to inspectable CUDA/PTX
and then through the standard NVIDIA toolchain to a GPU binary. Generated source
remains available as an escape hatch, while the default editing path keeps
hardware decisions in \cakeir{} where they remain analyzable. External numerical
comparison and on-device timing remain authoritative.

\FloatBarrier

\section{Analysis scope and validation}
\label{app:analysis-contract}
\label{app:soundness}

The harness evaluates typed schedules against the safety, conformance, semantic,
and data-consistency categories in Table~\ref{tab:analysis-categories}. For
supported constructs, it either accepts the candidate or returns a localized
finding. Missing information or analysis coverage is reported explicitly rather
than treated as a successful check.

Static analysis is a pre-compile gate only within its modeled domain; it does
not prove global GPU correctness or capture all microarchitectural behavior.
Coverage continues to expand, and both false positives and false negatives can
occur. 

\section{Production-kernel evolution trajectories}
\label{app:production-evolution}

These figures supplement the production endpoints in
Section~\ref{sec:eval-frontier}, using each source session's own objective and
denominator.

\subsection{KDA prefill}
\label{app:kda-evolution}

\begin{figure}[H]
  \centering
  \includegraphics[width=\textwidth]
    {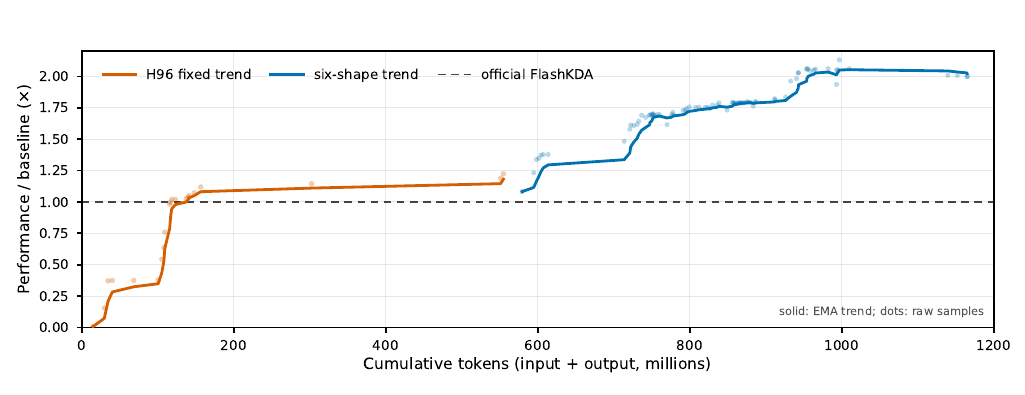}
  \caption{KDA prefill evolution on B200. Orange is fixed
  $H{=}96$, $S{=}8192$ bring-up; blue is six-shape geometric-mean speedup over
  official FlashKDA. All points pass correctness.}
  \label{fig:kda-prefill-evolution-token-performance}
\end{figure}

Figure~\ref{fig:kda-prefill-evolution-token-performance} separates fixed-shape
bring-up from the later six-shape campaign because the phases use different
metrics.

\subsection{TinyGEMM}
\label{app:tinygemm-evolution}

\begin{figure}[H]
  \centering
  \includegraphics[width=\textwidth]
    {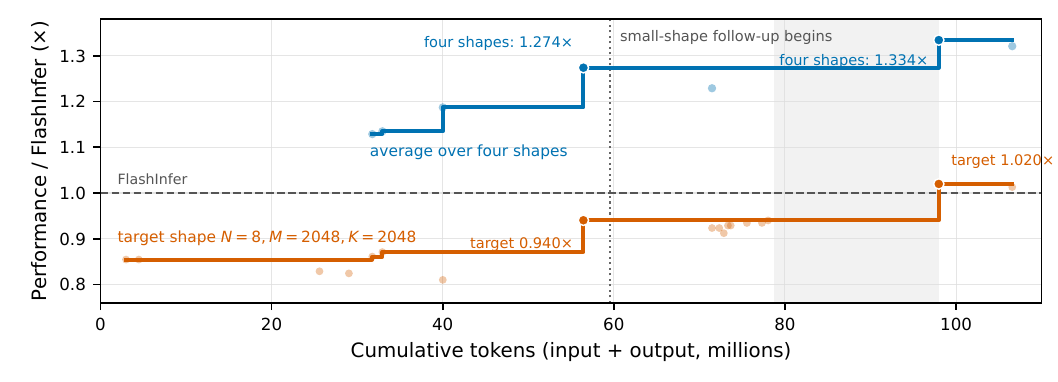}
  \caption{TinyGEMM evolution on B200. Orange tracks
  $N{=}8$, $M{=}2048$, $K{=}2048$; blue is the geometric mean over four
  recurring shapes, including orange. Dots are valid checkpoints, staircases
  are best-so-far, and the dotted line begins the follow-up.}
  \label{fig:tinygemm-evolution-token-performance}
\end{figure}

Figure~\ref{fig:tinygemm-evolution-token-performance} shows four-shape
evolution and a follow-up for the remaining small-shape regression. The
follow-up raises the target from $0.940\times$ to $1.020\times$ and the
four-shape mean from $1.274\times$ to $1.334\times$.

\subsection{Alpha-MoE}
\label{app:alphamoe-evolution}

\begin{figure}[H]
  \centering
  \includegraphics[width=\textwidth]
    {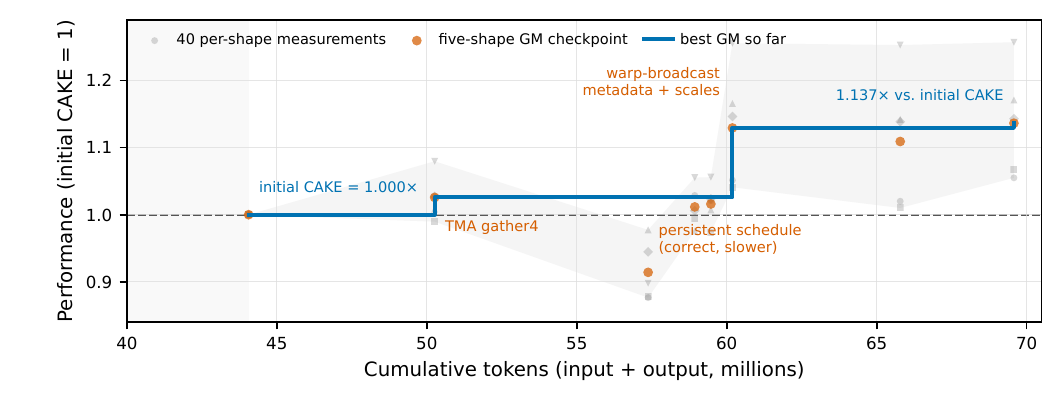}
  \caption{Alpha-MoE W8A8 Hopper-to-Blackwell rewrite on B200. Gray shows
  per-shape CUPTI medians; orange is the five-shape geometric mean (GM); blue is
  the best GM; shading is pre-checkpoint bring-up.}
  \label{fig:alphamoe-evolution-token-performance}
\end{figure}

Figure~\ref{fig:alphamoe-evolution-token-performance} normalizes five shapes to
their first correct \cake{} checkpoint. The final geometric mean is
$1.137\times$; this is an internal evolution metric, not a TensorRT-LLM
comparison.

\FloatBarrier

\section{Known-kernel reproduction details}
\label{app:known-kernel-details}

Table~\ref{tab:known-kernel-loc} gives the fixed shapes, relative performance,
and audited implementation sizes summarized in
Section~\ref{sec:eval-known}.

\begin{table}[H]
  \caption{Known-kernel reproduction for LLM-critical kernels. Relative
  performance is measured against the listed reference at the same B200 shape.
  \cakeir{} LOC counts the kernel IR; reference columns report audited device
  and supporting source where available. For DSv4, reference LOC is restricted
  to source reachable under the fixed S8 route and runtime constants.}
  \label{tab:known-kernel-loc}
  \centering
  \footnotesize
  \setlength{\tabcolsep}{4pt}
  \renewcommand{\arraystretch}{1.06}
  \begin{tabularx}{\linewidth}{@{}lYcrrrr@{}}
    \toprule
    & & & & \multicolumn{3}{c}{\thead{Physical LOC}} \\
    \cmidrule(lr){5-7}
    \thead{Family} & \thead{Variant} & \thead{Shape} & \thead{Rel. perf.} &
      \thead{\cakeir{}} &
      \shortstack{\thead{Ref.}\\[-1pt]\thead{dev.}} &
      \shortstack{\thead{Ref.+}\\[-1pt]\thead{support}} \\
    \midrule
    \rowtint
      & FWD, BF16 non-causal & S1
      & 1.0045 & 430 & 2369 & 3039 \\
    \rowtint
    \multirow{-2}{*}{FA4}
      & BWD, BF16 non-causal & S1
      & 1.0470 & 514 & 2552 & 3690 \\
    TRTLLM GQA
      & Decode, FP16 & S2
      & 1.043 & 783 & 8515 & 8515 \\
    \rowtint
      & 1D1D GEMM, FP8 & S3
      & 1.0370 & 221 & 516 & 657 \\
    \rowtint
      & \mbox{Grouped GEMM, BF16 masked} & S4
      & 1.0174 & 401 & 442 & 559 \\
    \rowtint
      & MQA indexer, FP8 & S5
      & 1.2700 & 480 & 704 & 972 \\
    \rowtint
      & MQA indexer, FP4 & S5
      & 1.2730 & 392 & 704 & 972 \\
    \rowtint
    \multirow{-5}{*}{DeepGEMM}
      & \mbox{Paged MQA indexer, FP4} & S6
      & 1.0036 & 395 & 779 & 1177 \\
    CUTLASS MLA
      & Decode, BF16, TMA & S7
      & 1.2174 & 845 & 1860 & 2449 \\
      & Decode, BF16 & S8
      & 1.1297 & 1299 & 13609 & 13790 \\
    \multirow{-2}{*}{\shortstack[l]{DSv4 sparse MLA}}
      & Decode, FP8 & S8
      & 0.9649 & 1393 & 8942 & 9125 \\
    \bottomrule
  \end{tabularx}\par
  \vspace{3pt}
  \parbox{\linewidth}{\scriptsize\raggedright
    \textit{Shape key.}
    S1: $B=4$, $H=32$, $S=8192$, $D=128$;
    S2: $B=128$, $QH=64$, $KVH=8$, $S_{\mathrm{kv}}=4096$, $D=128$, page 16;
    S3: $M=4096$, $N=7168$, $K=4096$;
    S4: 256 groups $\times$ $M=128$, $N=4096$, $K=7168$;
    S5: $H=32$, $D=128$, $S_q=1024$, $S_{\mathrm{kv}}=2048$;
    S6: $B=256$, $H=64$, $D=128$, avgKV 4096, block 64, next 1;
    S7: DeepSeek-V3, $B=128$, $S_{\mathrm{kv}}=4096$, page 128;
    S8: DeepSeek-V4 sparse MLA, $B=3$, ragged query
    lengths $[3,4,5]$, $H=128$, and $D_{qk}=D_v=512$.}
\end{table}

\FloatBarrier

\section{Dispatcher portfolio details}
\label{app:dispatcher-details}

\begin{figure}[H]
  \centering
  \definecolor{DispatchPanelFill}{HTML}{F7F9FC}
  \definecolor{DispatchPanelBorder}{HTML}{CBD5E1}
  \definecolor{DispatchGuardFill}{HTML}{FFF4E8}
  \definecolor{DispatchGuardBorder}{HTML}{D9A05B}
  \definecolor{DispatchPerf}{HTML}{596DC2}
  \definecolor{DispatchPerfFill}{HTML}{EEF2FF}
  \definecolor{DispatchText}{HTML}{1F2933}
  \resizebox{\textwidth}{!}{%
  \begin{tikzpicture}[font=\sffamily, text=DispatchText]
    % Panel backgrounds.
    \filldraw[fill=DispatchPanelFill, draw=DispatchPanelBorder,
      line width=0.7pt, rounded corners=3pt] (0.05,0.05) rectangle (7.85,8.70);
    \filldraw[fill=DispatchPanelFill, draw=DispatchPanelBorder,
      line width=0.7pt, rounded corners=3pt] (8.05,0.05) rectangle (16.50,8.70);

    % (a) Production KNN-build families.
    \node[anchor=west, font=\bfseries\small] at (0.40,8.30)
      {(a) KNN build};
    \filldraw[fill=DispatchGuardFill, draw=DispatchGuardBorder,
      line width=0.65pt, rounded corners=2pt] (0.42,7.28) rectangle (7.48,7.92);
    \node[font=\bfseries\scriptsize, align=center] at (3.95,7.60)
      {8 outer families (112 shapes total)\\
       80 distinct final routes observed};

    \node[anchor=west, font=\bfseries\scriptsize] at (0.42,6.96) {outer family};
    \node[font=\bfseries\tiny] at (3.95,6.96) {shapes/routes};
    \node[anchor=west, font=\bfseries\scriptsize] at (5.02,6.96)
      {$G_{\mathrm{span}}$};

    \foreach \y in {6.55,5.87,5.19,4.51,3.83,3.15,2.47,1.79} {
      \draw[draw=DispatchPanelBorder, line width=1.1pt] (4.92,\y) -- (6.55,\y);
    }

    \node[anchor=west, font=\scriptsize] at (0.42,6.55) {D64};
    \node[font=\scriptsize] at (3.95,6.55) {7 / 5};
    \draw[draw=DispatchPerf, line width=1.5pt] (4.92,6.55) -- (5.15,6.55);
    \fill[DispatchPerf] (5.15,6.55) circle (1.6pt);
    \node[anchor=west, font=\scriptsize] at (5.27,6.55) {$1.130\times$};

    \node[anchor=west, font=\scriptsize] at (0.42,5.87) {D128 low-K BF16};
    \node[font=\scriptsize] at (3.95,5.87) {38 / 17};
    \draw[draw=DispatchPerf, line width=1.5pt] (4.92,5.87) -- (5.38,5.87);
    \fill[DispatchPerf] (5.38,5.87) circle (1.6pt);
    \node[anchor=west, font=\scriptsize] at (5.50,5.87) {$1.258\times$};

    \node[anchor=west, font=\scriptsize] at (0.42,5.19) {D128 low-K FP16};
    \node[font=\scriptsize] at (3.95,5.19) {1 / 1};
    \draw[draw=DispatchPerf, line width=1.5pt] (4.92,5.19) -- (5.55,5.19);
    \fill[DispatchPerf] (5.55,5.19) circle (1.6pt);
    \node[anchor=west, font=\scriptsize] at (5.67,5.19) {$1.353\times$};

    \node[anchor=west, font=\scriptsize] at (0.42,4.51) {D128 mid-K};
    \node[font=\scriptsize] at (3.95,4.51) {32 / 30};
    \draw[draw=DispatchPerf, line width=1.5pt] (4.92,4.51) -- (5.68,4.51);
    \fill[DispatchPerf] (5.68,4.51) circle (1.6pt);
    \node[anchor=west, font=\scriptsize] at (5.80,4.51) {$1.425\times$};

    \node[anchor=west, font=\scriptsize] at (0.42,3.83) {D128 large-K};
    \node[font=\scriptsize] at (3.95,3.83) {10 / 5};
    \draw[draw=DispatchPerf, line width=1.5pt] (4.92,3.83) -- (6.55,3.83);
    \fill[DispatchPerf] (6.55,3.83) circle (1.6pt);
    \node[anchor=west, font=\scriptsize] at (6.67,3.83) {$1.909\times$};

    \node[anchor=west, font=\scriptsize] at (0.42,3.15) {D192};
    \node[font=\scriptsize] at (3.95,3.15) {1 / 1};
    \draw[draw=DispatchPerf, line width=1.5pt] (4.92,3.15) -- (5.69,3.15);
    \fill[DispatchPerf] (5.69,3.15) circle (1.6pt);
    \node[anchor=west, font=\scriptsize] at (5.81,3.15) {$1.432\times$};

    \node[anchor=west, font=\scriptsize] at (0.42,2.47) {D256};
    \node[font=\scriptsize] at (3.95,2.47) {8 / 8};
    \draw[draw=DispatchPerf, line width=1.5pt] (4.92,2.47) -- (5.61,2.47);
    \fill[DispatchPerf] (5.61,2.47) circle (1.6pt);
    \node[anchor=west, font=\scriptsize] at (5.73,2.47) {$1.386\times$};

    \node[anchor=west, font=\scriptsize] at (0.42,1.79) {high-D};
    \node[font=\scriptsize] at (3.95,1.79) {15 / 13};
    \draw[draw=DispatchPerf, line width=1.5pt] (4.92,1.79) -- (6.28,1.79);
    \fill[DispatchPerf] (6.28,1.79) circle (1.6pt);
    \node[anchor=west, font=\scriptsize] at (6.40,1.79) {$1.758\times$};

    \filldraw[fill=DispatchPerfFill, draw=DispatchPerf, line width=0.6pt,
      rounded corners=2pt] (0.42,0.46) rectangle (7.48,1.00);
    \node[font=\bfseries\scriptsize] at (3.95,0.73)
      {overall $G_{\mathrm{span}}=1.418\times$};

    % (b) Production Flash-KMeans routes.
    \node[anchor=west, font=\bfseries\small] at (8.38,8.30)
      {(b) Flash-KMeans};
    \filldraw[fill=DispatchGuardFill, draw=DispatchGuardBorder,
      line width=0.65pt, rounded corners=2pt] (8.38,7.28) rectangle (16.17,7.92);
    \node[font=\bfseries\scriptsize] at (12.275,7.60)
      {12 final routes (124 shapes total)};

    \node[anchor=west, font=\bfseries\scriptsize] at (8.38,6.96) {final route};
    \node[anchor=east, font=\bfseries\scriptsize] at (12.02,6.96) {shapes};
    \node[anchor=west, font=\bfseries\scriptsize] at (12.40,6.96)
      {$G_{\mathrm{span}}$};

    \foreach \y in {6.50,6.05,5.60,5.15,4.70,4.25,3.80,3.35,2.90,2.45,2.00,1.55} {
      \draw[draw=DispatchPanelBorder, line width=1.1pt] (12.45,\y) -- (14.80,\y);
    }

    \node[anchor=west, font=\scriptsize] at (8.38,6.50) {gap-fused general};
    \node[anchor=east, font=\scriptsize] at (12.02,6.50) {64};
    \draw[draw=DispatchPerf, line width=1.5pt] (12.45,6.50) -- (13.09,6.50);
    \fill[DispatchPerf] (13.09,6.50) circle (1.6pt);
    \node[anchor=west, font=\scriptsize] at (13.21,6.50) {$1.753\times$};

    \node[anchor=west, font=\scriptsize] at (8.38,6.05) {D288 split-K};
    \node[anchor=east, font=\scriptsize] at (12.02,6.05) {14};
    \draw[draw=DispatchPerf, line width=1.5pt] (12.45,6.05) -- (13.49,6.05);
    \fill[DispatchPerf] (13.49,6.05) circle (1.6pt);
    \node[anchor=west, font=\scriptsize] at (13.61,6.05) {$2.227\times$};

    \node[anchor=west, font=\scriptsize] at (8.38,5.60) {aligned-shape fallback};
    \node[anchor=east, font=\scriptsize] at (12.02,5.60) {12};
    \draw[draw=DispatchPerf, line width=1.5pt] (12.45,5.60) -- (12.82,5.60);
    \fill[DispatchPerf] (12.82,5.60) circle (1.6pt);
    \node[anchor=west, font=\scriptsize] at (12.94,5.60) {$1.439\times$};

    \node[anchor=west, font=\scriptsize] at (8.38,5.15) {D144--176 padded tail};
    \node[anchor=east, font=\scriptsize] at (12.02,5.15) {8};
    \draw[draw=DispatchPerf, line width=1.5pt] (12.45,5.15) -- (13.12,5.15);
    \fill[DispatchPerf] (13.12,5.15) circle (1.6pt);
    \node[anchor=west, font=\scriptsize] at (13.24,5.15) {$1.786\times$};

    \node[anchor=west, font=\scriptsize] at (8.38,4.70) {D112 specialized};
    \node[anchor=east, font=\scriptsize] at (12.02,4.70) {6};
    \draw[draw=DispatchPerf, line width=1.5pt] (12.45,4.70) -- (12.80,4.70);
    \fill[DispatchPerf] (12.80,4.70) circle (1.6pt);
    \node[anchor=west, font=\scriptsize] at (12.92,4.70) {$1.414\times$};

    \node[anchor=west, font=\scriptsize] at (8.38,4.25) {D352 split-K};
    \node[anchor=east, font=\scriptsize] at (12.02,4.25) {6};
    \draw[draw=DispatchPerf, line width=1.5pt] (12.45,4.25) -- (14.79,4.25);
    \fill[DispatchPerf] (14.79,4.25) circle (1.6pt);
    \node[anchor=west, font=\scriptsize] at (14.91,4.25) {$3.757\times$};

    \node[anchor=west, font=\scriptsize] at (8.38,3.80) {D480 split-K};
    \node[anchor=east, font=\scriptsize] at (12.02,3.80) {6};
    \draw[draw=DispatchPerf, line width=1.5pt] (12.45,3.80) -- (13.81,3.80);
    \fill[DispatchPerf] (13.81,3.80) circle (1.6pt);
    \node[anchor=west, font=\scriptsize] at (13.93,3.80) {$2.597\times$};

    \node[anchor=west, font=\scriptsize] at (8.38,3.35) {micro-D hybrid};
    \node[anchor=east, font=\scriptsize] at (12.02,3.35) {2};
    \draw[draw=DispatchPerf, line width=1.5pt] (12.45,3.35) -- (12.71,3.35);
    \fill[DispatchPerf] (12.71,3.35) circle (1.6pt);
    \node[anchor=west, font=\scriptsize] at (12.83,3.35) {$1.309\times$};

    \node[anchor=west, font=\scriptsize] at (8.38,2.90) {micro-D pipelined};
    \node[anchor=east, font=\scriptsize] at (12.02,2.90) {2};
    \draw[draw=DispatchPerf, line width=1.5pt] (12.45,2.90) -- (12.53,2.90);
    \fill[DispatchPerf] (12.53,2.90) circle (1.6pt);
    \node[anchor=west, font=\scriptsize] at (12.65,2.90) {$1.095\times$};

    \node[anchor=west, font=\scriptsize] at (8.38,2.45) {large paired};
    \node[anchor=east, font=\scriptsize] at (12.02,2.45) {2};
    \draw[draw=DispatchPerf, line width=1.5pt] (12.45,2.45) -- (12.63,2.45);
    \fill[DispatchPerf] (12.63,2.45) circle (1.6pt);
    \node[anchor=west, font=\scriptsize] at (12.75,2.45) {$1.217\times$};

    \node[anchor=west, font=\scriptsize] at (8.38,2.00) {D128 specialized};
    \node[anchor=east, font=\scriptsize] at (12.02,2.00) {1};
    \draw[draw=DispatchPerf, line width=1.5pt] (12.45,2.00) -- (12.48,2.00);
    \fill[DispatchPerf] (12.48,2.00) circle (1.6pt);
    \node[anchor=west, font=\scriptsize] at (12.60,2.00) {$1.036\times$};

    \node[anchor=west, font=\scriptsize] at (8.38,1.55) {D224 TMEM};
    \node[anchor=east, font=\scriptsize] at (12.02,1.55) {1};
    \draw[draw=DispatchPerf, line width=1.5pt] (12.45,1.55) -- (12.48,1.55);
    \fill[DispatchPerf] (12.48,1.55) circle (1.6pt);
    \node[anchor=west, font=\scriptsize] at (12.60,1.55) {$1.037\times$};

    \filldraw[fill=DispatchPerfFill, draw=DispatchPerf, line width=0.6pt,
      rounded corners=2pt] (8.38,0.46) rectangle (16.17,1.00);
    \node[font=\bfseries\scriptsize] at (12.275,0.73)
      {overall $G_{\mathrm{span}}=1.803\times$};
  \end{tikzpicture}%
  }
  \caption{GPU-span speedup by dispatch family (KNN build) and final route
  (Flash-KMeans). Values are per-row geometric means from same-session CUPTI
  measurements. Baselines are FlashLib 0.2.0 and our tuned FlashLib
  implementation, respectively.}
  \label{fig:dispatcher-portfolios}
\end{figure}
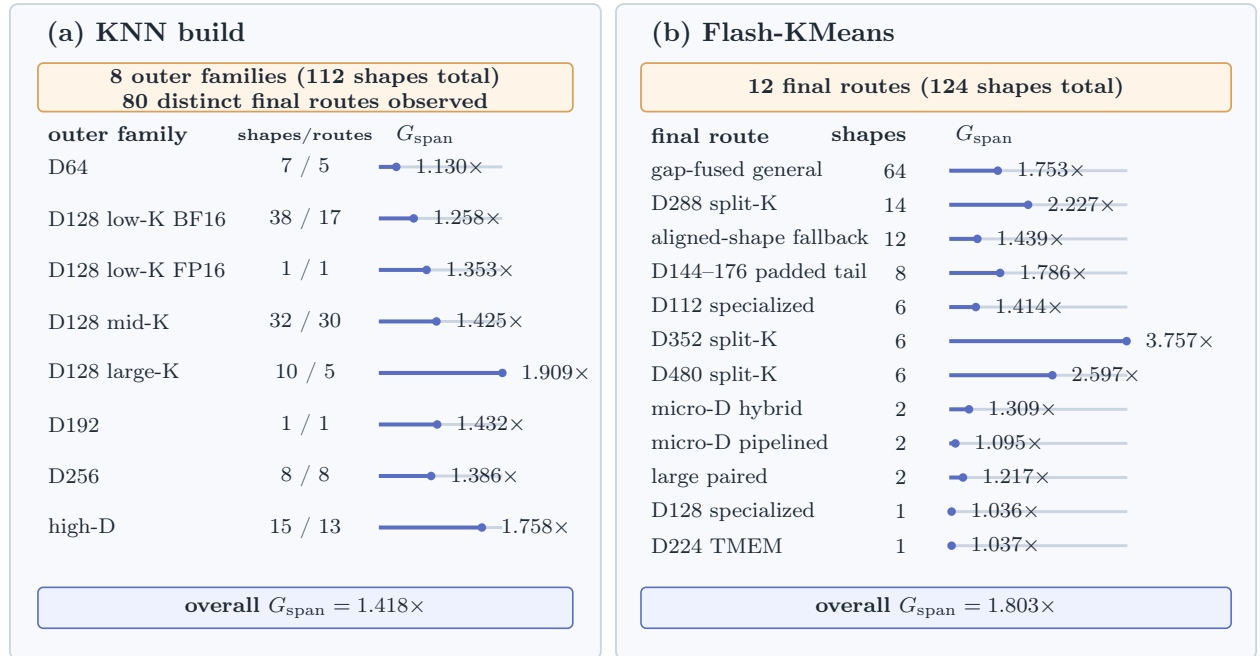

Figure~\ref{fig:dispatcher-portfolios} reports the route-level composition
behind the aggregate generalization results in
Section~\ref{sec:generalization}.

\FloatBarrier

\end{document}